\documentclass[letterpaper, journal]{IEEEtran}

\IEEEoverridecommandlockouts                              

\usepackage{amsmath,amsfonts}
\usepackage{algorithmic}
\usepackage{algorithm}
\usepackage{array}
\usepackage[caption=false,font=normalsize,labelfont=sf,textfont=sf]{subfig}
\usepackage{textcomp}
\usepackage{stfloats}
\usepackage{url}
\usepackage{verbatim}
\usepackage{graphicx}
\usepackage{cite}
\usepackage[T1]{fontenc}
\usepackage{amsthm}
\usepackage{amssymb}
\usepackage{bbm}
\usepackage{mathtools}
\usepackage{cite}
\usepackage{booktabs}
\usepackage{tabularx}
\usepackage[scientific-notation=false]{siunitx}
\usepackage{algorithm,algorithmic}
\usepackage{graphics}
\usepackage{tikz,pgfplots}
\usepackage{textcomp}
\usepackage{makecell}
\usepackage{fontawesome}

\usepackage{siunitx}
\usepackage{graphicx}
\usepackage{epstopdf}
\usepackage{calc}
\usetikzlibrary{positioning,chains}
\usetikzlibrary{shapes.geometric}
\usetikzlibrary{calc}
\usetikzlibrary{trees}
\usetikzlibrary{decorations.pathreplacing,calc,tikzmark}
\usetikzlibrary{backgrounds, fit}
\usetikzlibrary{positioning,arrows,petri,shapes.misc}
\usetikzlibrary{arrows.meta, bending, calc, fit, shapes, patterns}
\usepackage{pgfplots}
\usepackage{pgfplotstable}
\usepackage{xcolor}
\usepackage{import}
\usepackage{comment}

\usepackage{blindtext}
\usepackage{url}
\usepackage[hidelinks]{hyperref}
\usepackage[capitalise]{cleveref}
\crefname{section}{Sec.}{Sec.}
\Crefname{section}{Sec.}{Sec.}
\crefname{figure}{Fig.}{Fig.}
\Crefname{figure}{Fig.}{Fig.}
\crefname{equation}{}{}
\Crefname{equation}{}{}
\crefname{algorithm}{Alg.}{Alg.}
\Crefname{algorithm}{Alg.}{Alg.}
\usepackage{dirtree} 

\usepackage[nohyperlinks, nolist]{acronym}

\usepackage{pifont}
\newcommand{\cmark}{\ding{51}}%
\newcommand{\xmark}{\ding{55}}%

\usepackage{fancyvrb}

\DeclareMathOperator*{\ego}{\mathrm{ego}}
\DeclareMathOperator*{\obs}{\mathrm{obs}}

\DeclareMathOperator*{\traj}{\mathrm{traj}}

\DeclareMathOperator{\state}{\mathbf{s}}
\DeclareMathOperator{\position}{\mathbf{p}}
\newcommand{\xposition}{p_\mathrm{x}}
\newcommand{\yposition}{p_\mathrm{y}}
\newcommand{\orientation}{\theta}
\newcommand{\turingrate}{\omega}
\newcommand{\nvelocity}{v} 
\DeclareMathOperator{\nacceleration}{a} 
\newcommand{\xvelocity}{v_\mathrm{x}} 
\DeclareMathOperator{\xacceleration}{a_x} 
\newcommand{\yvelocity}{v_\mathrm{y}} 
\DeclareMathOperator{\yacceleration}{a_y}

\newcommand{\overtake}{\mathrm{overtake}}
\newcommand{\headon}{\mathrm{head\_on}}
\newcommand{\crossing}{\mathrm{crossing}}
\newcommand{\keep}{\mathrm{keep}}
\newcommand{\collisionpossible}{\mathrm{collision\_possible}}

\newtheorem{theorem}{Theorem}
\newtheorem{lemma}{Lemma}
\newtheorem{proposition}{Proposition}

\theoremstyle{definition}
\newtheorem{definition}{Definition}
\theoremstyle{definition}
\newtheorem{requirement}{COLREGS Requirement}

\definecolorset{RGB}{CODE}{}{%
    Gray, 140, 140, 140;
    Blue,   0,  51, 179%
}

\definecolorset{RGB}{TUM}{}{%
	Black,   0,   0,   0;
	Gray,   51,  51,  51;
	White, 0,  51, 179;
	Blue, 0, 101, 189;
	Ivory,  140, 140, 140;
	Orange, 0,  51, 179; 
	Green,  162, 173,   0;
	G,  140, 140, 140;
    B,   0,  51, 179%
}
\definecolor{TUMBlue}{RGB}{0, 101, 189}
\definecolor{TUMOrange}{RGB}{227, 114, 34}
\definecolor{TUMGreen}{RGB}{162, 173,   0}
\definecolor{TUMGray}{RGB}{51,  51,  51}
\definecolor{rebuttal}{RGB}{148,  0,  211}

\title{Provable Traffic Rule Compliance in \\ Safe Reinforcement Learning on the Open Sea}

\author{Hanna Krasowski and Matthias Althoff
\thanks{All authors are with Technical University of Munich, Germany; TUM School of Computation, Information and Technology, Department of Computer Engineering; Munich Center for Machine Learning (MCML).
        {\tt\small \newline \{hanna.krasowski, althoff\}@tum.de}}%
}

\begin{document}
\begin{acronym}
	\acro{colregs}[COLREGS]{Convention on the International Regulations for Preventing Collisions at Sea}
	\acro{rl}[RL]{Reinforcement learning}
	\acro{mpc}[MPC]{model predictive control}
\end{acronym}

\markboth{}%
{Krasowski \MakeLowercase{\textit{et al.}}: Provable Traffic Rule Compliance in Safe Reinforcement Learning}

\IEEEpubid{}

\maketitle
\begin{abstract}
    For safe operation, autonomous vehicles have to obey traffic rules that are set forth in legal documents formulated in natural language. Temporal logic is a suitable concept to formalize such traffic rules. Still, temporal logic rules often result in constraints that are hard to solve using optimization-based motion planners. \ac{rl} is a promising method to find motion plans for autonomous vehicles. However, vanilla \ac{rl} algorithms are based on random exploration and do not automatically comply with traffic rules. Our approach accomplishes guaranteed rule-compliance by integrating temporal logic specifications into \ac{rl}. Specifically, we consider the application of vessels on the open sea, which must adhere to the \ac{colregs}.
    To efficiently synthesize rule-compliant actions, we combine predicates based on set-based prediction with a statechart representing our formalized rules and their priorities.
    Action masking then restricts the \ac{rl} agent to this set of verified rule-compliant actions. In numerical evaluations on critical maritime traffic situations, our agent always complies with the formalized legal rules and never collides while achieving a high goal-reaching rate during training and deployment. In contrast, vanilla and traffic rule-informed \ac{rl} agents frequently violate traffic rules and collide even after training.
\end{abstract}
\begin{IEEEkeywords}
	Safe reinforcement learning, autonomous vessels, temporal logic, provable guarantees, collision avoidance.
\end{IEEEkeywords}

\acresetall
\section{Introduction} \label{ch:intro}
\ac{rl} has provided promising results for a variety of motion planning tasks, e.g., autonomous driving \cite{Kiran2021, Ye2021}, robotic manipulation \cite{El-Shamouty2020, Han2023-surveyRL}, and autonomous vessel navigation \cite{Sarhadi2022-survey, Heiberg2022, Xu2022}. \ac{rl} algorithms learn a capable policy through random exploration. As random exploration is inherently unsafe, \ac{rl} agents are mainly trained and tested in simulation only. To transfer the capabilities of \ac{rl}-based motion planning systems to the physical world, the agents have to be safe. 
Safe \ac{rl} extends \ac{rl} algorithms with safety considerations. Most safe \ac{rl} approaches constrain the learning softly, e.g., by integrating risk measures in the reward function or by adapting the optimization problem for obtaining a policy considering constraints \cite{krasowski2023provably}. However, for safety-critical tasks, such as motion planning in the physical world, hard safety guarantees are necessary, which most safe \ac{rl} approaches cannot provide.

Provably safe \ac{rl} achieves hard safety guarantees during training and operation by combining \ac{rl} with formal methods \cite{krasowski2023provably}. The safety specifications regarded in provably safe \ac{rl} are so far mainly \emph{avoid specifications}, i.e., it is ensured that unsafe areas and actions are always avoided. However, the notion of safety for real-world tasks is often more complex than avoiding unsafe sets. For autonomous vehicles, legal safety is usually required, meaning that vehicles do not cause collisions by obeying traffic rules \cite{Vanholme2013, Mehdipour2023}. 
To apply formal methods, these traffic rules need to be formalized. Temporal logic is suited to formalize traffic rules \cite{Vanholme2013,Tuncali2018, Vasile2017, Maierhofer.2020, Esterle2020, Krasowski.2021}, as it can capture their spatial and temporal dependencies well. Still, efficient and generalizable integration of formalized traffic rules in motion planning approaches is an open research problem.

\IEEEpubidadjcol

\begin{figure*}
	\vspace{0.2cm}
	\centering
	\resizebox{2\columnwidth}{!}{%
	\input{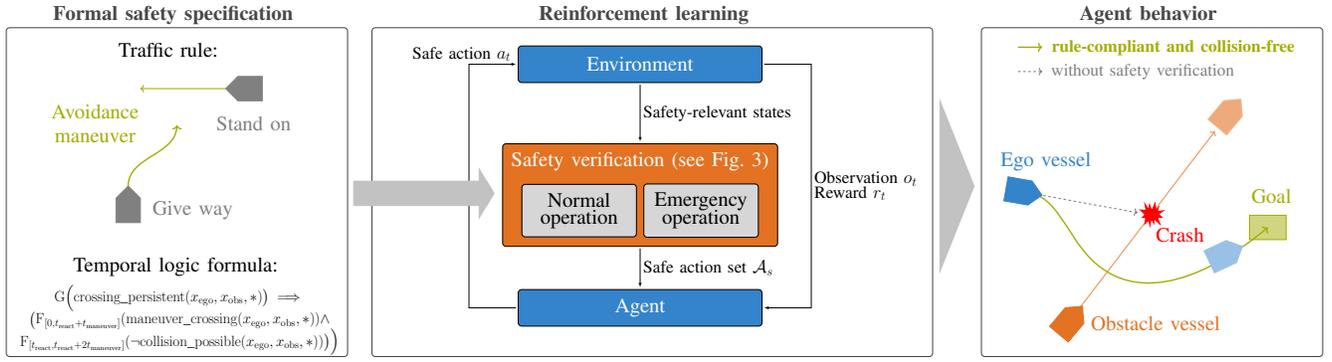}}
	\caption{Proposed provably safe \ac{rl} approach for autonomous vessels. First, traffic rules for collision avoidance are formalized with temporal logic (see \cref{ch:prelim}). Based on the formal specification, the set of rule-compliant actions is identified (see \cref{ch:online_verification} and \cref{ch:maneuver_synthesis}), which are integrated in the \ac{rl} process so that the agent can only select actions that are rule-compliant (see \cref{ch:reinforcement_learning}). Note that the statechart in \cref{fig:statechart} details the computation of verified rue-compliant actions and comprises two modes: normal operation and emergency operation. The resulting safe agent achieves rule compliance and collision avoidance during training and deployment, while agents without the safety verification of actions violate the formalized traffic rules and collide still after training (see \cref{ch:experiments}).}
	\label{fig:overview_paper}
\end{figure*}

In this work, we propose a provably safe \ac{rl} approach that ensures legal safety by complying with traffic rules formalized in temporal logic for the application of autonomous vessel navigation. \Cref{fig:overview_paper} displays the concept of our approach. We develop a statechart that reflects the formalized traffic rules and their hierarchy. Regular collision avoidance rules are followed as long as there is no immediate collision risk, and an emergency operation that executes a last-minute maneuver is immediately activated once a collision becomes likely. For the regular collision avoidance rules, an application-specific maneuver synthesis method based on a search algorithm is developed to efficiently identify actions that are compliant with traffic rules.
For emergency operation, we detect imminent collision of the vessels using set-based reachability analysis and design an emergency controller that aims to prevent collisions as much as possible. 
Rule-compliant actions for both regular and emergency operation are computed online based on our statechart and are used to constrain the \ac{rl} agent so it can only select verified actions. Our main contributions are:
\begin{itemize}
	\item We are the first to introduce a safe RL approach that ensures provable satisfaction of open-sea maritime collision avoidance rules, which are formally specified via temporal logic;
	\item We improve our previously formalized maritime traffic rules \cite{Krasowski.2021}, newly formalize the last-minute maneuver rule from the Convention on the \ac{colregs}, and develop a rule-compliant emergency controller;
	\item Our provably safe maneuver synthesis for discrete action spaces efficiently identifies safe actions online;
	\item We train provably safe \ac{rl} and safety-informed \ac{rl} agents on critical maritime traffic situations and evaluate their performance in different deployment configurations on handcrafted and recorded maritime traffic data.
\end{itemize}

The remainder of this article is structured as follows: We present and discuss related literature in \cref{ch:related}, introduce relevant concepts published preliminarily to this article and state the problem in \cref{ch:prelim}. We present the formalized traffic rules and prove that a statechart models the traffic rules in \cref{ch:online_verification}. We describe our rule-compliant maneuver synthesis in \cref{ch:maneuver_synthesis}. The 
\ac{rl} approach is detailed in \cref{ch:reinforcement_learning}. In \cref{ch:experiments}, we discuss our experimental results on critical maritime traffic situations and conclude in \cref{ch:conclusions}. 

\section{Related Work}\label{ch:related}
We  categorize related work into safety specifications for maritime motion planning, motion planning approaches for autonomous vessels, and provably safe \ac{rl}.

\paragraph{Safety specification for maritime motion planning}
The notion of safety in maritime motion planning is usually rule compliance with maritime traffic rules describing collision avoidance maneuvers \cite{Zhang2021}. The most relevant maritime traffic rules for collision avoidance are specified in the \ac{colregs} \cite{IMO.1972}. Often, these traffic rules are indirectly integrated in the motion planning approach, e.g., through geometric thresholds \cite{Kuwata.2014, Zhao.2019a, Guo.2020, Zhang.2019, Junmin.2020, He.2017}, virtual obstacles \cite{Chiang.2018}, or cost functions \cite{Benjamin.2004, Johansen.2016, Heiberg2022, Eriksen2019, Kufoalor2019, Stankiewicz2021}. However, these approaches usually do not capture the temporal properties of collision avoidance rules, and the implemented interpretation of the \ac{colregs} is often intransparent.  

Another concept is to formalize the traffic rules and directly use them in motion planning. This is a more faithful consideration of traffic rules than the previously mentioned indirect integration. Additionally, the rule formalization is usually parameterized, which eases adaptions.
Temporal logic is suited to formalize \ac{colregs} since it captures temporal dependencies and thus can model sophisticated specifications of encounter situations. There are two relevant studies that formalize maritime traffic rules with temporal logic.
Torben et al.~\cite{Torben2023} formalize \ac{colregs} with signal temporal logic for automatic testing of autonomous vessels. This has the advantage that robustness measures specified through signal temporal logic formulas can be used as costs for motion planning approaches, since they quantify rule compliance. 
Krasowski et al.~\cite{Krasowski.2021} formalize \ac{colregs} with metric temporal logic and evaluate their compliance on real-world maritime traffic data. They discuss that the \ac{colregs} are currently not well posed for more than two vessels, which needs to be addressed by regulators to make autonomous vessels admissible for commercial deployment in the real-world.
How to best employ temporal logic formalizations for motion planning approaches as presented by \cite{Torben2023, Krasowski.2021} is an open research question, for which we propose a solution in this work.

\paragraph{Motion planning for vessels}
The motion planning literature can be categorized into single-agent and multi-agent motion planning problems \cite{LiuBucknall2018}, where multi-agent settings are often distinguished into cooperative \cite{Wu2022, Wu2024, Zhao2021} and non-cooperative \cite{Zhang2023, Wen2022} settings. In this article, we regard single-agent motion planning. Maritime motion planners are often divided into three building blocks \cite{Fossen2011}: a guidance system generating reference trajectories, a control system for tracking reference trajectories, and a state observer\footnote{Often referred to as a navigation system in the maritime literature.}.
For example, one line of single-agent motion planning research employs search-based algorithms based on motion primitives, e.g., rapidly-exploring random trees \cite{Zhang2022, Stankiewicz2021, Enevoldsen2021}. Other studies employ \ac{mpc} \cite{Johansen.2016, Tsolakis2022,Kufoalor2019} to obtain an optimal control signal. In contrast to using search algorithms based on a finite amount of motion primitives, \ac{mpc} directly optimizes the controller in the continuous state and input space. In particular, the studies \cite{Johansen.2016, Kufoalor2019} show promising results on multi-obstacle scenarios and Kufoalor et al.~\cite{Kufoalor2019} even evaluate their approach in real-world experiments with two obstacle vessels.
However, for \ac{mpc} an optimization problem must be solved repeatedly, which can be computationally costly.

\ac{rl} is a well-suited machine learning approach to solve single-agent motion planning tasks in uncertain environments \cite{Meyer2020, Heiberg2022, Chun2021, Xu2022, Xie2024, Fan2023}. Regarded scenarios are usually on the open sea with other non-reactive dynamic obstacles \cite{Chun2021, Xu2022, Xie2024} and static obstacles \cite{Meyer2020, Heiberg2022, Fan2023}. 
To achieve a behavior that adheres to maritime traffic rules, the reward function considers rule compliance to minimize risks, but does not guarantee compliance because the reward function is only maximized \cite{Meyer2020, Heiberg2022, Chun2021, Xu2022, Xie2024, Fan2023}. In contrast, provably safe \ac{rl} approaches ensure safety \cite{krasowski2023provably}.

\paragraph{Provably safe RL}
Provably safe \ac{rl} approaches ensure safety during training and operations. There are three conceptual approaches for provably safe \ac{rl} \cite{krasowski2023provably}: action replacement, action projection, and action masking. In this article, we present an action masking approach, for which the agent can only choose actions that are verified as safe. Most research on action masking considers discrete action spaces; common applications are autonomous driving \cite{Fulton2018, Fulton2019, Mirchevska2018, krasowski2020safe, Brosowsky2021, Krasowski2022} and power systems \cite{Tabas2022}. Usually, the action verification is tailored to the specific application and, thus, cannot be directly transferred to other applications.

Another way to distinguish provably safe \ac{rl} approaches is by the safety specification. Most approaches consider safety specifications that can be formalized as containment in a safe set or avoiding intersection with unsafe sets. A few works regard safety specifications based on temporal logic \cite{Alshiekh2018, Konighofer2020, Li2019}, which can additionally model temporal dependencies in safety specifications. The studies \cite{Alshiekh2018, Konighofer2020} use model checking to determine whether a given action fulfills a linear temporal logic formula, which expresses the safety specification. Their approaches are transferable between applications but limited to discrete action and state spaces. In contrast, Li et al.~\cite{Li2019} leverage linear temporal logic specifications to synthesize control barrier functions, which are used to project unsafe actions proposed by the agent to safe actions. This allows them to apply their approach to continuous action and state spaces. However, their approach cannot deal with dynamic obstacles that are not controllable, such as other traffic participants. To the best of our knowledge, we are the first to formulate a provably safe \ac{rl} approach for the application of autonomous vessels and to include temporal safety specifications in the online safety verification of \ac{rl} agents while operating in a continuous state space.

\section{Preliminaries and Problem Statement}\label{ch:prelim}

\paragraph{Notation and dynamics}
We denote sets by calligraphic letters, vectors are boldfaced, and predicates are written in Roman typestyle. The Minkowski sum is defined as $\mathcal{Y}_1 \oplus \mathcal{Y}_2 = \{y_1 + y_2  \ |\  y_1 \in \mathcal{Y}_1, y_2 \in \mathcal{Y}_2\}$ and the set-based multiplication is defined as $\mathcal{Y}_1 \mathcal{Y}_2 = \{y_1 y_2 | y_1 \in \mathcal{Y}_1, y_2 \in \mathcal{Y}_2\}$.  A traffic rulebook $ \left\langle  \Phi, \leq \right\rangle$ is a tuple where $\Phi$ is the set of formalized rules and $\leq$ is the order \cite{Censi2019}. We denote that the model $\Xi$ and its initial state $\xi$ entail  the rulebook $\left\langle  \Phi, \leq \right\rangle$ by $\, \Xi, \xi \models \left\langle  \Phi, \leq \right\rangle$.

The state of a vessel $\state \in \mathbb{R}^4$ consists of the position $\position = [\xposition, \yposition] \in \mathbb{R}^2$ in the Cartesian coordinate frame as well as the orientation $\orientation \in \mathbb{R}$, and the orientation-aligned velocity $\nvelocity \in \mathbb{R}$. The operator $\mathtt{proj}_{\square}$ projects a state to the state dimensions indicated by $\square$ and $\mathrm{R}(\Upsilon) = \{ \mathtt{R}(\upsilon)| \upsilon \in \Upsilon\}$ denotes the set of rotation matrices for the angles $\Upsilon$ with $ \mathtt{R}(\upsilon)$ being the rotation matrix for the angle $\upsilon$. To model the ego vessel (i.e., the autonomous vessel we control), we use a yaw-constrained model $\Omega_\mathrm{yc}$ with orientation-aligned acceleration $\nacceleration \in \mathbb{R}$ and turning rate $\turingrate \in \mathbb{R}$ as control inputs:
\begin{equation} \label{eq:yaw_model}
	\dot{\state} = \begin{bmatrix}
		 \dot{\xposition} \\
		 \dot{\yposition} \\
		 \dot{\orientation} \\
		 \dot{\nvelocity}
	\end{bmatrix} = 
\begin{bmatrix}
	\cos (\orientation) \, \nvelocity \\
	\sin (\orientation) \, \nvelocity \\
	\turingrate \\
	\nacceleration
\end{bmatrix} .
\end{equation}
The control input is denoted as $\mathbf{u} (t) = [\nacceleration (t), \turingrate (t)]$ and the initial state as $\state_0$.

\paragraph{Set-based prediction of vessels}
To obtain predictions that enclose all possible behaviors of a traffic participant, the concept of set-based predictions for road traffic participants \cite{Koschi2021} can be transferred to maritime traffic. The fundamental idea is to define abstract models and perform reachability analysis for them. We first specify the dynamics used for the prediction, and then introduce the reachable sets and occupancy sets. Finally, we discuss the special case of a closed-loop system. 

For vessels, we assume that the abstract model is a point-mass model $\Omega_\mathrm{pm}$ with velocity and acceleration constraints:
\begin{align}
	\dot{\xposition}(t) = \xvelocity(t), & \; \dot{\yposition}(t)  = \yvelocity(t),  \label{eq:set_based_prediction_model} \\
	\dot{\xvelocity}(t)  = \xacceleration(t), &\; \dot{\yvelocity}(t)  = \yacceleration(t), \notag \\ 
	\text{subject to} \qquad & \sqrt{ \xvelocity(t)^2 + \yvelocity(t)^2} \leq \nvelocity_\mathrm{pm,max}  \notag \\ 
	& \sqrt{ \xacceleration(t)^2 + \yacceleration(t)^2} \leq \nacceleration_\mathrm{pm,max}. \notag
\end{align}
The maximum velocity and maximum acceleration are denoted by $\nacceleration_\mathrm{pm,max}$ and $\nvelocity_\mathrm{pm,max}$, respectively. 
To ensure formal safety of our approach, the two constraints must be chosen such that the point-mass model over-approximates the behavior of vessels using reachset conformance \cite{Roehm2019}. The state of the model $\Omega_\mathrm{pm}$ is abbreviated by $\mathbf{x} = [\xposition, \yposition, \xvelocity, \yvelocity]$. 

The time-point reachable sets for the model $\Omega_\mathrm{pm}$ are calculated with set-based reachability analysis \cite{Althoff2010diss} based on the initial state $\state_0$, time step size $\Delta t$, and the time horizon $t_\mathrm{pred}$. Note that the state $\state_0$ is transformed into $\mathbf{x}_0$ by using trigonometry to convert $[\nvelocity, \orientation]$ into $[\xvelocity, \yvelocity]$.
The time-interval reachable sets are computed as in \cite{Althoff2010diss, Wetzlinger2023TAC} and are denoted by $\mathcal{R}_{\Delta t}(\state_0, \Omega_\mathrm{pm}, t_\mathrm{pred})$. To obtain the occupancy sets from the time-interval reachable sets, the reachable sets are projected to the position domain and enlarged by the spatial extensions of the vessel $\mathcal{V}$ rotated by all possible reachable orientations using the Minkowski sum:
\begin{align}
	\mathcal{O}_{\mathrm{pm}} &(\state_0, \Omega_\mathrm{pm}, t_\mathrm{pred}, \mathcal{V}) = \label{eq:set_based_occupancy} \\  &\mathtt{proj}_{\position} \left(\mathcal{R}_{\Delta t}(\state_0,  \Omega_\mathrm{pm}, t_\mathrm{pred})\right) \oplus \notag \\ &\Big( \mathrm{R}\left(\mathtt{proj}_{\orientation} \left(\mathcal{R}_{\Delta t}(\state_0,  \Omega_\mathrm{pm}, t_\mathrm{pred})\right)\right) \, \mathcal{V} \Big).\notag
\end{align}
For a detailed derivation of the occupancy sets, we refer the interested reader to \cite[Sec.~V-A]{Koschi2021}.

The occupancy sets $\mathcal{O}_{\mathrm{pm}}$ are calculated for the open-loop system $\Omega_\mathrm{pm}$ since we do not have access to the control input of other traffic participants. However, for an ego vessel, we have a precise model $\Omega_\mathrm{yc}$ and access to the control input. Thus, the forward simulation of our closed-loop system with the control input $\mathbf{u}(t)$ provides the time-point reachable sets. 
The occupancy is denoted by: 
\begin{align}
	\label{eq:trajectory_occupany}
	\mathcal{O}_{\traj} (\state_0, \Omega_\mathrm{yc}, t_\mathrm{pred}, \mathcal{V}, \mathbf{u}(t)). 
\end{align}

\paragraph{Problem statement}\label{ch:problem_statement}
The \ac{colregs} specify the traffic rules for collision avoidance on the open sea for power-driven vessels in natural language. These traffic rules are satisfiable for two vessels. For more than two vessels, unsatisfiable traffic situations can occur, e.g., a vessel needs to keep its course and speed with respect to one vessel and perform an avoidance maneuver with respect to another vessel. The \ac{colregs} do not specify how to adequately resolve such conflicting situations with more than two vessels. Due to the lack of legal specifications, we regard traffic situations with two vessels only. In particular, we assume:
\begin{enumerate}
	\item The traffic situation is an open-sea situation without traffic signs, traffic separation zones, or static obstacles;
	\item There is one traffic participant vessel $obs$ and one autonomous vessel $ego$, which are both power-driven;
	\item The dynamics of the autonomous vessel is modeled by \cref{eq:yaw_model};
	\item The current state of the traffic participant vessel $\state_{\obs}$ is observed without measurement errors;
	\item In the initial state of the traffic situation, none of the collision avoidance rules specified in the \ac{colregs} apply.
\end{enumerate}

We define the traffic rulebook $ \left\langle  \Phi, \leq \right\rangle$ that describes the legally relevant collision avoidance rules of the \ac{colregs} given our assumptions 1) and 2). The formal traffic rules are denoted by $\Phi$ and the hierarchy between them by $\leq$. Based on the traffic rules, we search for an \ac{rl} approach, which ensures that the \ac{rl} agent only selects safe, i.e., rule-compliant, actions leading to rule-compliant trajectories.  
Thus, the overall problem is to find
\begin{align}
	\pi_s \colon \mathcal{S} &\,  \rightarrow \mathcal{A}_s \label{eq:problemstatement}\\
	 \text{where} & \quad \zeta_{\pi_s} \models \left\langle  \Phi, \leq \right\rangle \notag.
\end{align}
The observation space of the \ac{rl} agent is $\mathcal{S}$, the set of provably rule-compliant actions is $\mathcal{A}_s$, and the trajectories $\zeta_{\pi_s}$ are solutions of \cref{eq:yaw_model} when following the \ac{rl} policy $\pi_s$.
To address this problem, we first introduce the rulebook $\left\langle  \Phi, \leq \right\rangle$, and prove that a statechart $\Gamma$ entails the rulebook in \cref{ch:online_verification}. Then, we describe the synthesis of rule-compliant maneuvers and detail the safe-by-design action selection in \cref{ch:maneuver_synthesis}. Finally, we describe the \ac{rl} specification in \cref{ch:reinforcement_learning}.

\section{Specification}\label{ch:online_verification}
Our previous work \cite{Krasowski.2021} formalizes the \ac{colregs} rules specifying collision avoidance between two power-driven vessels on the open sea. 
The temporal operators used are $\mathrm{G}$,  $\mathrm{F}$, and $\mathrm{U}$, and if there is a subscript, the temporal operator is evaluated over the time interval indicated by the subscript. The operator $\mathrm{G}(\phi)$ evaluates to true iff $\phi$ is true for all future time steps. In contrast, for the operator $\mathrm{F}(\phi)$, $\phi$ only has to be true for at least one future time step. The until operator $\phi_1 \mathrm{U} \phi_2$ is true iff $\phi_1$ holds true for all time steps until $\phi_2$ holds true. In this section, we introduce the legal specification through a rulebook and detail the novel formalization of the emergency rule. Finally, we introduce the statechart $\Gamma$ and show that it models the specification.

\subsection{Traffic Rulebook}

\begin{table*}[tb]
	\vspace{0.2cm}
	\centering
	\caption{Overview formalized marine traffic rules integrated in the safety verification}
	\renewcommand{\arraystretch}{2}
	\begin{tabular*}{2\columnwidth}{lcc}
		\toprule
		\textbf{Rule} &  \textbf{Temporal logic formula} \\
		\midrule
		$R_1^{\ddagger}$& $\mathrm{G(is\_emergency(\state_{\ego},\state_{\obs}, *) \implies (emergency\_maneuver \; \mathrm{U} \,  is\_ emergency\_resolved(\state_{\ego},\state_{\obs}, *) ))}$      \\ 
		$R_2$& $\mathrm{G\Big(safe\_speed}(\state_{ego}, \nvelocity_\mathrm{max})\Big)$     \\[0.2cm]
		$R_3^{\dagger}$ & \makecell[c]{$\mathrm{G\Big(persistent\_crossing}(\state_{\ego},\state_{\obs}, *)  \implies$ \\ $\big( \mathrm{F}_{[0,t_\mathrm{react} + t_\mathrm{maneuver}]} (\mathrm{maneuver\_crossing}(\state_{\ego}, \state_{\obs}, *)) \land 
			\mathrm{F}_{[t_\mathrm{react},t_\mathrm{react}+ 2 t_\mathrm{maneuver}]} (\lnot \mathrm{collision\_possible}(\state_{\ego}, \state_{\obs}, t_\mathrm{horizon}^\mathrm{check}))\big)\Big)$} \\[0.4cm]
		$R_4^{\dagger}$   & \makecell[c]{$\mathrm{G\Big(persistent\_head\_on}(\state_{\ego},\state_{\obs}, *) \implies$ \\ $\big( \mathrm{F}_{[0,t_\mathrm{react} + t_\mathrm{maneuver}]} (\mathrm{maneuver\_head\_on}(\state_{\ego}, \state_{\obs}, *)) \land  \mathrm{F}_{[t_\mathrm{react},t_\mathrm{react}+ 2 t_\mathrm{maneuver}]} (\lnot \mathrm{collision\_possible}(\state_{\ego}, \state_{\obs}, t_\mathrm{horizon}^\mathrm{check}))\big)\Big)$} \\[0.4cm]
		$R_5^{\dagger}$  & \makecell[c]{$\mathrm{G\Big(persistent\_overtake}(\state_{\ego},\state_{\obs}, *) \implies$ \\ $\big( \mathrm{F}_{[0,t_\mathrm{react} + t_\mathrm{maneuver}]} (\mathrm{maneuver\_overtake}(\state_{\ego}, \state_{\obs}, *)) \land  \mathrm{F}_{[t_\mathrm{react},t_\mathrm{react}+ 2 t_\mathrm{maneuver}]} (\lnot \mathrm{collision\_possible}(\state_{\ego}, \state_{\obs}, t_\mathrm{horizon}^\mathrm{check})) \big)\Big)$ }  \\[0.4cm]
		$R_6$  & $\mathrm{G\Big(keep}(\state_{\ego},\state_{\obs}, *) \implies \big(\mathrm{no\_turning}(\state_{\ego}, *) \mathrm{U} \, \lnot \mathrm{keep}(\state_{\ego}, \state_{\obs}, *)\big)\Big)$      \\
		\bottomrule
	\end{tabular*}
	\begin{center}
		{\footnotesize \textit{Note: Additional arguments are abbreviated by $*$, rules adapted from \cite{Krasowski.2021} are marked with $\dagger$, and new rules are marked with $\ddagger$.}}
	\end{center}
	\vspace{-0.5cm}
	\label{tab:rules}
\end{table*}

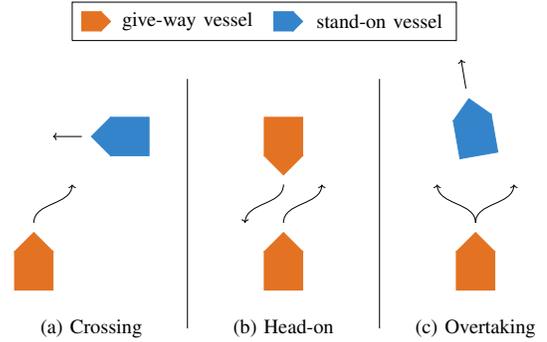
\begin{figure}[tb]
	\centering
	\resizebox{0.8\columnwidth}{!}{%
		\def\vessel#1#2#3#4{
	\begin{scope}[shift={#1}, rotate=#2, scale=#4]
		\draw [fill=#3,draw=#3](-1,1) -- (0,2) -- (1,1) -- cycle;
		\draw [fill=#3,draw=#3](-1,-1) rectangle (1,1);
	\end{scope}
}
\begin{tikzpicture}[auto, scale=0.8]  
	\vessel{(7,7)}{90}{TUMBlue!80}{1}
	\vessel{(2,0)}{0}{TUMOrange}{1}
	\vessel{(25,7)}{10}{TUMBlue!80}{1}
	\vessel{(25,0)}{0}{TUMOrange}{1}
	\vessel{(15,0)}{0}{TUMOrange}{1}
	\vessel{(15,7)}{180}{TUMOrange}{1}
	\draw[very thick] (10,-3) -- (10,10);
	\draw[very thick] (20,-3) -- (20,10);
	\draw(5,-3) node {\Huge{(a) Crossing}};
	\draw(25,-3) node {\Huge{(c) Overtaking}};
	\draw(15,-3) node {\Huge{(b) Head-on}};
	\draw[->, very thick] (4.5,7) -- (3,7);
	\draw[->, very thick] (24.5,9.5) -- (24.25,11);
	\draw (2,2.5) edge[out=90,in=-90,->, very thick] (4,4.5);
	\draw (25,2.5) edge[out=90,in=-90,->, very thick] (23,4.5);
	\draw (25,2.5) edge[out=90,in=-90,->, very thick] (27,4.5);
	\draw (15,2.5) edge[out=90,in=-90,->, very thick] (17,4.5);
	\draw (15,4.5) edge[out=-90,in=90,->, very thick] (13,2.5);
	\begin{scope}[xshift=-10cm]
	\vessel{(15,13)}{-90}{TUMOrange}{0.5}
	\vessel{(25,13)}{-90}{TUMBlue!80}{0.5}
	\draw(30,13) node {\Huge{stand-on vessel}};
	\draw(20,13) node {\Huge{give-way vessel}};
	\draw[very thick] (14,12) -- (14,14) -- (34,14) -- (34,12) -- (14,12);
	\end{scope}
\end{tikzpicture}
	}
	\caption{Encounter situations and rule-compliant maneuvers specified in the \ac{colregs} (adapted from \cite{Krasowski.2021}).}
	\label{fig:encounter_situations}
\end{figure}

\Cref{tab:rules} lists all formalized rules considered in this work. While the predicates can be evaluated on any two vessels, the predicate arguments are set to be evaluated for the ego vessel with respect to an obstacle vessel according to the \ac{colregs}. The traffic rule $R_2$ enforces a safe speed, which is trivially ensured through the ego vessel dynamics. Thus, we do not include this rule in the traffic rulebook.

\begin{definition}[Rules $\Phi$]\label{def:rulebook} 
	The rulebook consist of rules $R_1$ and ${R_3-R_6}$ specified in \cref{tab:rules}.
\end{definition}

We introduce the emergency rule $R_1$ to reflect the \ac{colregs} specification that if the other vessel does not take appropriate actions for collision avoidance, the ego vessel has to react and perform a last-minute maneuver for collision risk minimization. 

\begin{requirement}[Rulebook order $\leq$]\label{def:rule_hierarchy} 
	Rule $R_1$ is always prioritized over rules $R_3$ - $R_5$, and $R_6$ has the lowest priority. Rules $R_3$ - $R_5$ are all of equal priority.
\end{requirement}

The predicates of rule $R_1$ are detailed in \cref{sec:emergency_predicates}. Note that we use the emergency maneuver to describe the last-minute maneuver, through which the ego vessel minimizes the collision risk and thereby achieves legal safety. Yet, in the literature, the term failsafe planning is also frequently used \cite{Magdici2016,krasowski2023provably}. 

Rules $R_3$ - $R_6$ describe how vessels have to behave in a \ac{colregs} encounter situation. In these encounter situations, the vessels are on a collision course meaning that the vessels would collide in the near future if no appropriate collision avoidance measures are taken. There are three different encounter situations specified in the \ac{colregs} as illustrated in \cref{fig:encounter_situations}: overtaking ($R_5$, $R_6$), crossing ($R_3$, $R_6$), and head-on encounters ($R_4$). In an encounter, a vessel can be a give-way or a stand-on vessel. A give-way vessel is required to change course and perform a collision avoidance maneuver. A stand-on vessel has the obligation to keep its course and speed. 
The predicate for determining a stand-on vessel is $\keep$ (see Appendix\ref{appendix:predicates}).
The stand-on rule $R_6$ has the lowest priority since whenever the other vessel changes its course so that the ego vessel becomes the give-way vessel, the give-way rules $R_3$ to $R_5$ are applied (see \cref{def:rule_hierarchy}).

To formalize that a give-way encounter is persistent for at least the reaction time, we use the following temporal logic specification, where $\mathrm{\{give\_way\}}$ can take the values from $\{ \crossing$, $\headon$, $\overtake \}$  (see Appendix\ref{appendix:predicates}) and $*$ denotes additional arguments for the predicates:
\begin{align*}
	&\mathrm{persistent\_\{give\_way\}} (\state_{\ego}, \state_{\obs}, *) = \\
	& \quad \lnot \mathrm{\{give\_way\}}(\state_{\ego}, \state_{\obs}, *) \, \land \\
	& \quad \mathrm{G_{[\Delta t, t_\mathrm{react}]} (\{give\_way\}} (\state_{\ego}, \state_{\obs}, *)).
\end{align*}
We assume that both vessels keep their course and speed to obtain rule-compliant predictions for their future states. These predicted states allow us to evaluate ahead of time if the encounter situation will persist long enough so that the ego vessel has to perform a collision avoidance maneuver. The reaction time $t_\mathrm{react}$ does not indicate the minimum required reaction time of a human operator but instead specifies how much time the human operator would require to decide if the encounter situation persists. Given a give-way encounter is detected, a rule-compliant collision avoidance maneuver has to be conducted until $\lnot \collisionpossible$ evaluates to true (see \cref{tab:rules} $R_3$ - $R_5$). The time interval for performing a rule-compliant maneuver is $t_\mathrm{react} + 2 t_\mathrm{maneuver}$, where $2 t_\mathrm{maneuver}$ approximates the time required for the maneuvering. 

\begin{requirement}[Maneuvering priority]\label{def:maneuver_hierarchy}
	Given a rule $R_i$ for $i \in \{3,...,5\}$ applies, rules $R_j$ for $j \neq i \land j \in \{3,...,5\}$ are not applied until $\lnot \collisionpossible$ is true.
\end{requirement}


\subsection{Emergency Rule Predicates}\label{sec:emergency_predicates} 
We use the predicate $\collisionpossible$
to determine if two vessels are on a collision course for rules $R_3$ - $R_6$. Because the rules $R_3$ - $R_6$ assume a constant velocity, we use the velocity obstacle concept \cite{Fiorini.1998} for this predicate.
However, the velocity obstacle concept is not sufficient for detecting imminent risk as necessary for $R_1$.  Thus, we present four predicates in this section that are relevant for our formalization of rule $R_1$.

First, we define an auxiliary position predicate determining if vessel $m$ is in a relative orientation sector of vessel $l$: 
\begin{align*}
	&\mathrm{in\_sector}(\state_l, \state_m, \underline{\beta},  \overline{\beta}) \iff \\
	& \quad \mathbf{h}_{\underline{\beta}}^T \mathtt{proj}_{\position} (\state_{m}) - b_{l,\underline{\beta}} \leq 0 \; \land \\
	& \quad \mathbf{h}_{\overline{\beta}}^T \mathtt{proj}_{\position} (\state_{m}) - b_{l,\overline{\beta}} > 0,
\end{align*}
where the lower relative orientation is $\underline{\beta}$ and the upper relative orientation is $\overline{\beta}$ relative to the orientation of vessel $l$. The normal vector $\mathbf{h}_i$ is the unit vector in the direction $i - \pi/2$ and $b_{l,i}$ is the offset to the origin for a line through the position of vessel $l$ in the direction $i$. We illustrate the sector predicate with two specific usages in \cref{fig:emergency_modes}.

Second, we use set-based prediction for rule $R_1$ to detect potential collisions in the near future.
In particular, we predict the future occupancy of the obstacle vessel until the time horizon $t_\mathrm{pred}$ as described in \cref{eq:set_based_occupancy} and that of the ego vessel as in \cref{eq:trajectory_occupany}, for the control sequence $\mathbf{u}_\mathrm{keep}(t) = [\SI{0}{\meter \per \second \squared}, \SI{0}{\radian \per \second}] $ to keep course and speed as demanded for stand-on vessels. 
If the ego occupancy and the predicted occupancy of the obstacle vessel intersect, the ego vessel is in an emergency situation:
\begin{align*}
	&\mathrm{is\_emergency} (\state_{\ego}, \state_{\obs}, \mathcal{V}_{\ego}, \mathcal{V}_{\obs}, t_\mathrm{pred},\mathbf{u}_\mathrm{keep}(t)) \iff \\
	&\quad \exists t \in [t_0, t_0 + t_\mathrm{pred}] : \mathcal{O}_{\mathrm{pm}}(\state_0, \Omega_\mathrm{pm}, t, \mathcal{V}_{\obs}) \, \cap \\
	& \quad \mathcal{O}_{\traj} (\state_{\ego}, \Omega_\mathrm{yc}, t, \mathcal{V}_{\ego}, \mathbf{u}_\mathrm{keep}(t)) \neq \emptyset,
\end{align*}
where $t_0$ is the current time.

Third, the predicate $\mathrm{emergency\_maneuver}$ describes a maneuver that minimizes the risk of collision for the specific traffic situation. 
We detail our interpretation of $\mathrm{emergency\_maneuver}$ in \cref{sec:emergency_controller}.

Fourth, an emergency situation is resolved when the obstacle vessel is behind the ego vessel, is moving away from the ego vessel, and the Euclidean distance between both is larger than a specified minimum distance $d_\mathrm{resolved}$:
\begin{align*}
	&\mathrm{is\_emergency\_resolved} (\state_{\ego}, \state_{\obs}, d_\mathrm{resolved}) \iff \\
	& \quad \underbrace{\mathrm{in\_sector}(\state_{\ego}, \state_{\obs}, 3\pi/2, \pi/2)}_{\textbf{obstacle is behind}} \, \land \\
	& \quad \underbrace{\mathtt{unit\_v}(\state_{\obs})^{T} \, \mathtt{unit\_v}(\state_{\ego}) \leq 0}_{\textbf{moving away}} \, \land \\
	&  \quad \underbrace{\| \mathtt{proj}_{\position} (\state_{\obs}) - \mathtt{proj}_{\position}(\state_{\ego})\|_2 \leq d_\mathrm{resolved}}_{\textbf{distance between vessels is large enough}}
\end{align*}
where the unit orientation vector of a state is ${\mathtt{unit\_v}(\state) = [\cos(\mathtt{proj}_{\orientation}(\state)),  \sin(\mathtt{proj}_{\orientation}(\state))]}$.


\subsection{Specification-compliant Statechart}

\begin{figure}
	\centering
	\resizebox{1\columnwidth}{!}{%
		\begin{tikzpicture}[auto,
	node distance = 23mm and 10mm,
	myarrow/.style={
		->,
		thick,
		shorten <=2pt,
		shorten >=2pt,}
	]
	\tikzstyle{state}=[draw, rounded corners=1.5mm,
	inner ysep=2mm, inner xsep=4mm,
	minimum height=20mm, align=left, minimum width= 3cm, font=\Large]
	\node[state, draw=TUMGreen, fill=TUMGreen!20] (noconflict)   {$\rho_0$ \\ No conflict};
	\node[state, draw=TUMBlue, fill=TUMBlue!20, node distance=1.2cm]  (standon) [below= of noconflict]{$\rho_1$ \\ Stand-on \\ (\cref{alg:rule_compliant_actions})};
	\node[state, draw=TUMBlue, fill=TUMBlue!20]  (crossing) [above= of noconflict]{$\rho_3$ \\ Crossing \\ (\cref{alg:rule_compliant_actions})};
	\node[state, draw=TUMBlue, fill=TUMBlue!20]  (overtake)  [right=3cm of crossing]          {$\rho_4$ \\Overtake \\  (\cref{alg:rule_compliant_actions})};
	\node[state, draw=TUMBlue, fill=TUMBlue!20]  (headon)  [left=3cm of crossing]          {$\rho_2$ \\ Head-on \\ (\cref{alg:rule_compliant_actions})};
	\node[state, draw=TUMOrange, fill=TUMOrange!20, node distance=1.5cm]  (emergency)   [above=of crossing]          { $\rho_5$ \\ Emergency operation \\ (\cref{alg:emergencymaneuvering})};
	\node[rounded corners=1.5mm,minimum width=1cm,inner sep=2mm,above right,draw=black,align=left,text width=7mm,fit=(noconflict)(standon)(crossing)(overtake)(headon), ultra thick] (Op) {};
	\node[above right=2mm and 4mm of Op.south west, ultra thick, text=black]   (temp2) {\Large{Normal operation}};
	
	\coordinate[below left= 10mm and 10mm of noconflict.south west]   (temp1);
	\path[{Circle[length=2mm,flex]}-{Latex[flex]}, bend left]
	(temp1) edge (noconflict.south west);
	\coordinate[above right=10mm and 10mm of headon.north west]   (temp4);
	\path[{Circle[length=2mm,flex]}-{Latex[flex]}, bend right]
	(temp4) edge (Op.north west);
	
	\coordinate (standonn1) at ($ (standon.north) + (-0.5,0) $);
	\draw [myarrow, -latex] (standonn1)  -- (standonn1|-noconflict.south) node [near start, yshift=0.4cm] {\large{$\lnot \mathrm{keep}$}};
	\coordinate (noconflicts1) at ($ (noconflict.south) + (0.5,0) $);
	\draw [myarrow, -latex] (noconflicts1)  -- (noconflicts1|-standon.north) node [near start] {\large{$\mathrm{keep}$}};
	\coordinate (standone1) at ($ (standon.east) + (0,-0.5) $);
	\coordinate (standone2) at ($ (standon.east) + (0,0.5) $);
	\coordinate (standone3) at ($ (standon.east) + (0.5,0) $);
	\coordinate (standone4) at ($ (standon.east) + (1.0,0) $);
	\draw (standone1)  edge[out=0,in=270,thick] (standone3);
	\draw (standone3)  edge[out=90,in=0,-latex, thick] (standone2);
	\node at (standone4) (legend1){\large{$\mathrm{keep}$}};
	\coordinate (noconflictn1) at ($ (noconflict.north) + (-0.25,0) $);
	\draw [myarrow, -latex] (noconflictn1)  -- (noconflictn1|-crossing.south) node [near end, yshift=-0.2cm] {\large{$\mathrm{persistent\_crossing}$}};
	\coordinate (crossings1) at ($ (crossing.south) + (0.0,0) $);
	\draw [myarrow, -latex] (crossings1)  -- (crossings1|-noconflict.north) node [near start, yshift=-0.2cm] {\large{$\lnot \mathrm{collision\_possible}$}};
	\coordinate (checke1) at ($ (Op.north) + (-0.5,0.15) $);
	\coordinate (checke2) at ($ (Op.north) + (0.5,0.15) $);
	\coordinate (emergency1) at ($ (emergency.south) + (0.5,0) $);
	\draw [myarrow, -latex] (checke1)  -- ( checke1|-emergency.south) node [near end] {\large{$\mathrm{is\_emergency}$}};
	\draw [myarrow, -latex] (emergency1)  -- (checke2) node [near end] {\large{$\mathrm{emergency\_resolved}$}};
	
	\draw [myarrow, -latex] (overtake.south)  -- (noconflict.north) node [near start,above, xshift=-1.2cm] {\large{$\lnot \mathrm{collision\_possible}$}};
	\coordinate (checkn2) at ($ (noconflict.north) + (1,0) $);
	\coordinate (overtakes2) at ($ (overtake.south) + (1,0) $);
	\draw [myarrow, -latex] (checkn2)  -- (overtakes2) node [near end, below, yshift=-0.8cm, xshift=-0.8cm] {\large{$\mathrm{persistent\_overtake}$}};
	
	\coordinate (headons2) at ($ (headon.south) + (-1.25,0) $);
	\coordinate (noconflictn2) at ($ (noconflict.north) + (-1.25,0) $);
	\draw [myarrow, -latex] (headons2)  -- (noconflictn2) node [near start,above, xshift=2.3cm] {\large{$\mathrm{persistent\_head\_on}$}};
	\coordinate (headons3) at ($ (headon.south) + (-0.25,0) $);
	\draw [-latex] (noconflictn1)  -- (headons3) node [near end, below,yshift=-0.8cm, xshift=-0.5cm] {\large{$\lnot \mathrm{collision\_possible}$}};
\end{tikzpicture}}
	\caption{Statechart $\Gamma$ modeling the legal safety specification with predicates at the transitions. The states for the regular collision avoidance rules $R_3$ - $R_6$ are depicted in blue and the emergency operation state for rule $R_1$ in red. For safety verification of actions, the algorithms identifying the set of rule-compliant actions (indicated in brackets) are employed given the current state $\rho_i$ of the statechart.}
	\label{fig:statechart}
\end{figure}
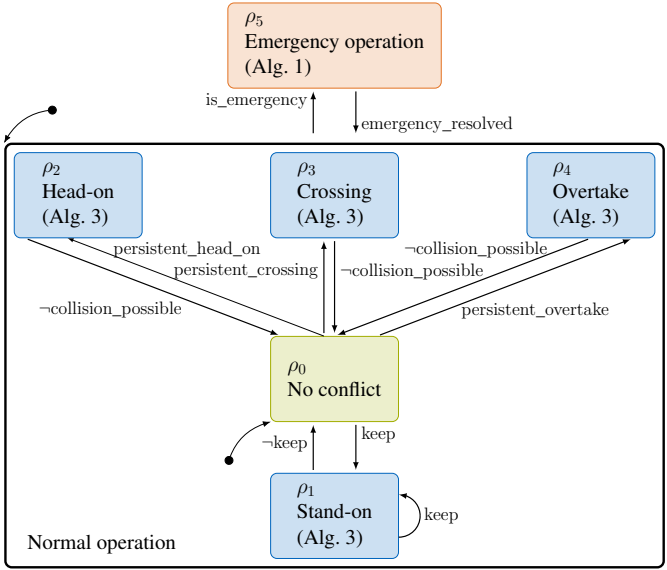

The overall rule specification is modeled by the statechart $\Gamma$ in \cref{fig:statechart}. Due to assumption 5), the initial state in every traffic situation is the state $\rho_0$.
There are two main states for normal operation and emergency operation. During normal operation, whenever the predicate $\collisionpossible$ is true, the corresponding maneuver state for $R_3$ - $R_6$ (see blue states in \cref{fig:statechart}) is entered and the collision avoidance maneuver is started.

\begin{proposition}\label{proposition:collision_possible_true} 
	For the states $\rho_i, \; \forall i \in \{1,...,4\}$, the predicate $\collisionpossible$ is true.
\end{proposition}
\begin{IEEEproof}
	This follows directly from the definition of the predicates $\keep$, $\headon$, $\crossing$, and $\overtake$ (see Appendix\ref{appendix:predicates}), which are true for the states of the statechart $\rho_1 - \rho_4$, respectively. 
\end{IEEEproof}

\begin{lemma}\label{lemma:mutal_exclusive} 
	For two specific vessels, at most one of the predicates $\keep$, $\headon$, $\crossing$, or $\overtake$ can be true at the same time.
\end{lemma}
\begin{IEEEproof}
	The predicates $\keep$, $\headon$, $\crossing$, and $\overtake$ cannot apply at the same time due to their mutually exclusive specification. The detailed proof is in Appendix\ref{appendix:lemma}.
\end{IEEEproof}

If an emergency situation is detected, the statechart transitions to the emergency operation state until the emergency situation is resolved. 

\begin{theorem}\label{theorem:statechart}
	It holds that $\Gamma, \rho_0 \models \left\langle  \Phi, \leq \right\rangle$ for the statechart $\Gamma$, its initial state $\rho_0$, and the rulebook $\left\langle  \Phi, \leq \right\rangle$. 
\end{theorem}

\begin{IEEEproof}
	The initial state $\rho_0$ fulfills the rulebook by assumption 5) (see \cref{ch:prelim}.c). We continue proving the compliance with each rule:	
	\smallskip
	
	\emph{(I) $R_1$:} \enspace If $\mathrm{is\_emergency}$ is true, $R_1$ applies and $R_3$ - $R_6$ do not (see \cref{def:rule_hierarchy}), which is realized by transitioning to $\rho_5$ (see \cref{fig:statechart}). 
	The state $\rho_5$ can only be exited iff $\mathrm{is\_emergency\_resolved}$ evaluates to true. Thus, the transition to and from $\rho_5$ directly represents $R_1$.\smallskip
	
	If $\collisionpossible \land \lnot \mathrm{is\_emergency}$ is true, then $\Gamma$ has to represent rules $R_3$ - $R_6$. Whenever $\collisionpossible$ becomes true, it can be deduced from  \cref{lemma:mutal_exclusive} and \cref{proposition:collision_possible_true} that the statechart transitions to a state $\rho_i$, $i \in \{1,...,4\}$.\smallskip
	
	\emph{(II) $R_3$ - $R_5$:} \enspace Based on \cref{def:maneuver_hierarchy}, once a rule $R_3$ - $R_5$ applies, i.e., the statechart is in either of the states $\rho_i$, $i \in \{2,...,4\}$, the respective avoidance maneuver has to be conducted until $\lnot \collisionpossible \lor \mathrm{is\_emergency}$ is true. For $\mathrm{is\_emergency}$, we showed in case (I) of this proof that $\Gamma$ models $ \left\langle  \Phi, \leq \right\rangle$. For $\lnot \collisionpossible $, the statechart $\Gamma$ transitions to $\rho_0$.\smallskip
	
	\emph{(III) $R_6$:} \enspace Once rule $R_6$ applies, i.e., $\keep$ is true, the statechart transitions to $\rho_1$ and stays there until $\lnot \keep \lor \lnot \collisionpossible \lor \mathrm{is\_emergency}$. If $\lnot \keep \land \collisionpossible$, an encounter of higher priority is present (see \cref{def:rule_hierarchy}) and $R_3$ - $R_5$ apply. In this situation, the statechart transitions to the states $\rho_i$ for $i \in \{2,...,4\}$ and the remaining proof steps are stated in case (III). Identically to case (III), if $\lnot \collisionpossible$ is true, the statechart $\Gamma$ transitions to $\rho_0$ and if $\mathrm{is\_emergency}$ the statechart transitions to $\rho_5$.
\end{IEEEproof}

\section{Rule-Compliant Maneuver Synthesis}\label{ch:maneuver_synthesis}
Given our specification-compliant statechart $\Gamma$, we need to identify rule-compliant actions for the individual states $\rho_i$ of the statechart. Trivially, for the state $\rho_0$ all actions are rule-compliant since no rules apply. We introduce the synthesis of emergency maneuvers in \cref{sec:emergency_controller} and of encounter maneuvers in \cref{sec:encounter verification}. Finally, we detail how we ensure a selection of only safe actions for the \ac{rl} agent in \cref{sec:safe_by_design}.

\begin{figure}[tb]
	\centering{
		\resizebox{1.0\columnwidth}{!}{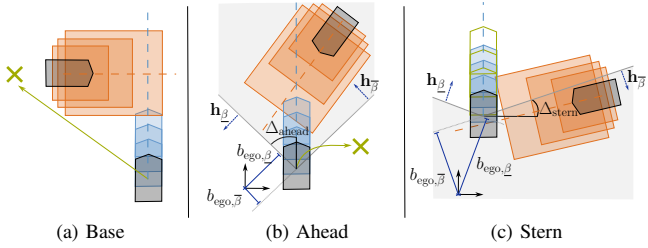}}
	\vspace{-0.6cm}
	\caption{Emergency controller modes with set-based occupancy prediction of obstacle vessel in orange and the occupancy of the ego vessel in blue for several time intervals. The orientation of the ego vessel and the obstacle vessel are indicated with dashed lines and emergency maneuver is depicted by green arrows or occupancies. The green cross indicates the target position for the base and ahead modes. The sectors, for which the predicate $\mathrm{in\_sector}$ is true, are shown in gray for the ahead and stern mode. The visualization of the sectors includes the arguments of predicate $\mathrm{in\_sector}$ in dark blue and the point of origin in black.}
	\label{fig:emergency_modes}
\end{figure}

\subsection{Emergency Maneuver}\label{sec:emergency_controller}
Once we detect an emergency situation, i.e., the statechart is in $\rho_5$, the ego vessel is legally required to evade the obstacle vessel in a manner that minimizes the risk of collision. 
In similar motion planning applications, such as autonomous driving \cite{Mehdipour2023}, autonomous aerial traffic \cite{Schouwenaars2004}, or human-robot environments \cite{Bouraine2012}, states that are safe for infinite time are used to identify a legally safe emergency maneuver. 
In contrast, the current \ac{colregs} do not state specifically how to interpret ``minimize risk'' or the characteristics of an invariably safe state. Thus, we cannot provide a formal specification. Consequently, we cannot verify risk minimizing behavior.
Nevertheless, we identify three situations in which different emergency maneuvers are appropriate: base mode, ahead mode, and stern mode (see \cref{fig:emergency_modes}).

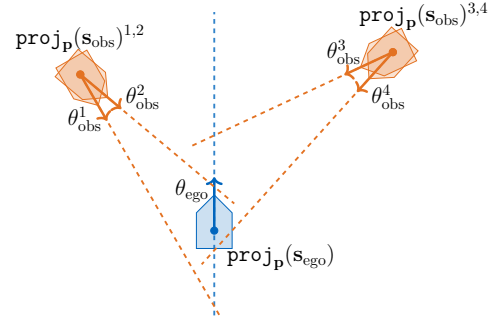
\begin{figure}[tb]
	\vspace{0.2cm}
	\centering
	\resizebox{0.75\columnwidth}{!}{%
		\def\vessel#1#2#3#4{
	\begin{scope}[shift={#1}, rotate=#2, scale=#4]
		\draw [fill=#3,draw=#3, fill opacity=0.2](-1,1) -- (0,2) -- (1,1) -- (1,-1) -- (-1,-1) -- cycle;
		\draw [fill=#3,draw=#3,->, ultra thick](0,0) -- (0,3);
		\draw [fill=#3,draw=#3](0,0) circle (0.2);
	\end{scope}
}
\begin{tikzpicture}[auto]  
	\draw[dashed, very thick, draw=TUMBlue!80] (0,-3.4) -- (0,3.4);
	\vessel{(0,-1.5)}{0}{TUMBlue}{0.4}
	\vessel{(-3,2)}{210}{TUMOrange}{0.4}
	\draw[dashed, very thick, rotate around={210:(-3,2)}, draw=TUMOrange] (-3,2) -- (-3,8.25);
	\vessel{(-3,2)}{230}{TUMOrange}{0.4}
	\draw[dashed, very thick, rotate around={230:(-3,2)}, draw=TUMOrange] (-3,2) -- (-3,6.5);
	\vessel{(4,2.5)}{115}{TUMOrange}{0.4}
	\draw[dashed, very thick, rotate around={115:(4,2.5)}, draw=TUMOrange] (4,2.5) -- (4,7.5);
	\vessel{(4,2.5)}{138}{TUMOrange}{0.4}
	\draw[dashed, very thick, rotate around={138:(4,2.5)}, draw=TUMOrange] (4,2.5) -- (4,8.9);
	\draw(1.5,-2.1) node {\scalebox{1}{\Large $\mathtt{proj}_{\position}(\state_{\ego})$}};
	\draw(-0.5,-0.6) node {\scalebox{1}{\Large $\orientation_{\ego}$}};
	\draw(-3,2.8) node {\scalebox{1}{\Large $\mathtt{proj}_{\position}(\state_{\obs})^{1,2}$}};
	\draw(-2.9,1.0) node {\scalebox{1}{\Large $\orientation_{\obs}^1$}};
	\draw(-1.6,1.5) node {\scalebox{1}{\Large $\orientation_{\obs}^2$}};
	\draw(4.7,3.3) node {\scalebox{1}{\Large $\mathtt{proj}_{\position}(\state_{\obs})^{3,4}$}};
	\draw(2.9,2.5) node {\scalebox{1}{\Large $\orientation_{\obs}^3$}};
	\draw(3.9,1.5) node {\scalebox{1}{\Large $\orientation_{\obs}^4$}};
\end{tikzpicture}
	}
	\caption{Visualization of turning direction cases. The obstacle vessel is depicted in orange and the ego vessel in blue. Arrows indicate orientations and positions are marked with dots. The turning direction case is indicated by the superscript. For cases 1 and 3, the ego vessel should turn right and for the cases 2 and 4, the ego vessel should turn left.}
	\label{fig:turning_direction}
\end{figure}

In the ahead case (see \cref{fig:emergency_modes}b), the obstacle vessel is in the ahead sector in front of the ego vessel, and the orientation difference between the ego vessel orientation and the reversed orientation of the obstacle vessel is at most $\Delta_\mathrm{ahead}$. This can be formalized as:
\begin{align}
	& \mathrm{ahead\_emergency} (\state_{\ego}, \state_{\obs}, \Delta_\mathrm{ahead}) \iff \label{eq:ahead_emergency}\\
	& \quad \mathtt{in}(\rho_5) \, \land \, \lnot \mathrm{orientation\_delta}(\state_{\ego}, \state_{\obs}, \Delta_\mathrm{ahead}, \pi) \, \land \notag\\
	& \quad  \mathrm{in\_sector}(\state_{\ego}, \state_{\obs}, -\Delta_\mathrm{ahead}, \Delta_\mathrm{ahead}), \notag
\end{align} 
where the predicate $\mathtt{in}(\rho_5)$ evaluates to true if and only if the statechart $\Gamma$ is in base state $\rho_5$.
In this ahead situation, steering to the stern of the obstacle vessel would lead to an even more critical situation, as both vessels would encounter each other head-on, given the obstacle vessel approximately keeps its speed and course. Thus, we instead require the ego vessel to turn $\SI{90}{\degree}$. The direction of turning is determined as presented in \cref{fig:turning_direction}. Depending on the situation, turning $\SI{90}{\degree}$ can be enough to resolve the emergency situations. Yet, if the emergency is not resolved and the traveled distance of the ego vessel from the start of the maneuver is larger than $d_\mathrm{min, ahead}$, the emergency controller switches to the base mode (see \cref{fig:emergency_modes}a) and steers the ego vessel behind the stern of the obstacle vessel.

The stern case is necessary for situations where the obstacle vessel is almost astern of the ego vessel and still relatively far away (see \cref{fig:emergency_modes}c):
\begin{align}
	&\mathrm{stern\_emergency} (\state_{\ego}, \state_{\obs}, \Delta_\mathrm{stern}, \mathbf{u}_\mathrm{acc}(t), \mathcal{V}_{\ego}, \mathcal{V}_{\obs},  \label{eq:stern_emergency} \\
	& \quad \quad \quad \quad \quad \quad \quad \quad t_\mathrm{pred}) \iff \notag \\
	& \quad \mathtt{in}(\rho_5) \, \land \notag \\
	& \quad \mathrm{in\_sector}(\state_{\ego}, \state_{\obs}, 3\pi/2 + \Delta_\mathrm{stern}, \pi/2 + \Delta_\mathrm{stern}) \, \land \notag \\
	& \quad \lnot \mathrm{is\_emergency} (\state_{\ego}, \state_{\obs}, \mathcal{V}_{\ego}, \mathcal{V}_{\obs}, t_\mathrm{pred},\mathbf{u}_\mathrm{acc}(t)), \notag
\end{align}
with the control sequence $\mathbf{u}_\mathrm{acc}(t) = [\nacceleration_\mathrm{stern}, \SI{0}{\radian\per\second}], \; \forall t \leq t_\mathrm{react}$ and then $\SI{0}{\meter\per\second \squared}, \SI{0}{\radian\per\second}], \; \forall t \leq t_\mathrm{react} < t \leq t_\mathrm{pred}$. By using the set-based prediction within this predicate, we ensure that we only use this controller mode if it is certain that accelerating would resolve the situation. In such a situation, performing an emergency maneuver that navigates the ego vessel to the stern of the obstacle vessel would be an unnecessarily long detour, given that a short acceleration period would also resolve the emergency situation.

For the base case (see \cref{fig:emergency_modes}a), the emergency situation can be safely resolved by steering to a position behind the stern of the obstacle vessel. The base emergency situation is formalized by:
\begin{align*}
	\mathrm{base}&\mathrm{\_emergency} \iff \\
	& \mathtt{in}(\rho_5) \, \land \, \lnot \mathrm{ahead\_emergency} \land \lnot \mathrm{stern\_emergency} \, \land  \\
	& \lnot \mathrm{is\_emergency\_resolved}.
\end{align*}
\Cref{alg:emergencymaneuvering} summarizes the control mode selection when entering the emergency operation state (see \cref{fig:statechart}) and is an instantiation of the predicate $\mathrm{emergency\_maneuver}$ of rule $R_1$ in \cref{tab:rules} for our problem statement. For base and ahead modes, the target positions are depicted in \cref{fig:emergency_modes} and obtained with the functions $\mathtt{get\_target\_ahead}$ and $\mathtt{get\_target\_base}$, respectively. Given the target position, a reachable desired position given the current state is identified and a control input toward this desired position is generated (for details on the controller design see Appendix\ref{appendix:controller}). The controller is abbreviated by the function $\mathtt{tracking\_controller}$.

\begin{algorithm}[tb]
	\caption{$\mathrm{emergency\_maneuver}(\state_{\ego},\state_{\obs}, *)$}
	\label{alg:emergencymaneuvering}
	\begin{algorithmic}[1]
		\renewcommand{\algorithmicrequire}{\textbf{Input:}}
		\renewcommand{\algorithmicensure}{\textbf{Output:}}
		\REQUIRE current state of ego vessel $\state_{\ego}$, current state of obstacle vessel $\state_{\obs}$, emergency mode $mode$, initial time $t_0$, time step size $\Delta t$, acceleration control sequence $\mathbf{u}_\mathrm{acc}(t)$  \\
		\ENSURE control input $\mathbf{u}(t_i)$ \\
		\STATE $\state_{\ego,0} = \mathtt{proj}_{\position}(\state_{\ego}), \state_{\obs,0} = \mathtt{proj}_{\position}(\state_{\obs}), t_i = t_0$ \\
		\WHILE{$\lnot \mathrm{emergency\_resolved}$}
		\IF{$\|\mathtt{proj}_{\position}(\state_{\ego,0}) - \mathtt{proj}_{\position}(\state_{\ego}) \|_2 > d_\mathrm{min,ahead}  \, \land mode = \text{ahead} $}
		\STATE $mode \leftarrow \text{base}$
		\ENDIF
		\IF{$mode = \text{stern}$}
		\STATE $ \nacceleration, \turingrate \leftarrow \mathbf{u}_\mathrm{acc}(t_i)$ 
		\ELSIF{$mode = \text{ahead}$}
		\STATE $\position_\mathrm{target} \leftarrow \mathtt{get\_target\_ahead}(\state_{\ego}, \state_{\ego,0}, \state_{\obs,0})$
		\STATE $\nacceleration, \turingrate \leftarrow \mathtt{tracking\_controller}(\state_{\ego},\position_\mathrm{target})$ 
		\ELSE
		\STATE $\position_\mathrm{target} \leftarrow \mathtt{get\_target\_base}(\state_{\ego}, \state_{\obs})$
		\STATE $ \nacceleration, \turingrate \leftarrow \mathtt{tracking\_controller}(\state_{\ego},\position_\mathrm{target})$ 
		\ENDIF
		\RETURN $\mathbf{u}(t_i) = [\nacceleration, \turingrate]$
		\STATE $ \state_{\ego}, \state_{\obs}\leftarrow \mathtt{step\_environment}(\nacceleration, \turingrate)$
		\STATE $t_i \leftarrow t_i + \Delta t$
		\ENDWHILE
	\end{algorithmic}
\end{algorithm} 

\subsection{Encounter Maneuvers}\label{sec:encounter verification}
Given a persistent give-way encounter is detected (i.e., the statechart in \cref{fig:statechart} transitions to one of the respective blue states $\rho_1, \ldots, \rho_4$), we identify safe actions that result in safe maneuvers resolving the encounter. 

Set-based predictions are well suited to verify that no collisions can occur if not all vessels comply with the regular collision avoidance rules $R_3$ - $R_6$. 
Still, for the regular collision avoidance rules, the implicit assumption in the \ac{colregs} is that both vessels comply with them. Thus, for identifying actions of the ego vessel that are rule-compliant with these rules, we can use a rule-compliant prediction for the obstacle vessel. 
For the three encounter situations specified (see \cref{fig:encounter_situations}), we differentiate between the ego vessel being the give-way ($R_3$ - $R_5$ apply) and stand-on vessel ($R_6$ applies). First, we detail the verification of actions given the ego vessel is the stand-on vessel, i.e., $\mathtt{in}(\rho_1)$. Then, we describe the more intricate synthesis given that the ego vessel is the give-way vessel ($\rho_i$ where $i \in \{2,...,4\}$), and finally, summarize our encounter action synthesis.

\paragraph{Stand-on maneuver synthesis for $\rho_1$} 
The trivial action for the predicate $\mathrm{keep}$ is $a_\mathrm{keep} = [\nacceleration = \SI{0}{\meter\per \second \squared}, \turingrate = \SI{0}{\radian\per \second}]$, i.e., keeping course and speed. Note that for this trivial action there is no explicit maneuver time and the action space needs to be restricted to this action until the ego vessel is not the stand-on vessel anymore or an emergency is detected (see \cref{fig:statechart}). 

\paragraph{Give-way maneuver synthesis for $\rho_2 - \rho_4$} 
For all give-way maneuvers, a significant change of orientation (i.e., $\Delta_\mathrm{large\_turn}$) is required so that other traffic participants can identify give-way maneuvers (see \cref{fig:encounter_situations}). For head-on and crossing encounters, the give-way vessel is always obliged to turn toward the right. For the overtake encounter, the suited turning direction depends on the orientation of the obstacle vessel, but this is not further specified in the \ac{colregs}. For our maneuver synthesis, the turning direction is to the left if the orientation of the obstacle vessel is more to the right than the orientation of the ego vessel, and otherwise  turning direction is to the right. 

Given the turning direction, we identify candidate actions, construct maneuvers based on them, and verify if a maneuver complies with the rules. Candidate actions lead to trajectories that already fulfill the minimal turning requirement within the maneuver segment time $t_\mathrm{m}$. A maneuver is verified if the predicate $\mathrm{collision\_possible}$ is false at the end of the maneuver and the occupancies of both vessels do not intersect during the maneuver:
\begin{align}
	& \mathrm{maneuver\_verified} (\mathbf{u}_\mathrm{m}(t), \state_{\ego}, \state_{\obs}, t_\mathrm{horizon}^\mathrm{check},  \mathbf{u}_\mathrm{keep}(t), \label{eq:predicate_maneuver_verified}\\
	& \quad \quad \quad \quad \quad\quad \quad \quad \; \mathcal{V}_{\ego},\mathcal{V}_\mathrm{{\obs+}}) \iff \notag \\
	& \quad \lnot \mathrm{collision\_possible} (\state_{\ego, t_\mathrm{end}}, \state_{\obs, t_\mathrm{end}}, t_\mathrm{horizon}^\mathrm{check}) \, \land \notag \\
	& \quad \quad \forall t \in [t_0, t_0 + t_\mathrm{end}] : \mathcal{O}_{\traj} (\state_{\ego}, \Omega_\mathrm{yc}, t, \mathcal{V}_{\ego}, \mathbf{u}_\mathrm{m}(t)) \, \cap \notag \\ 
	& \quad \quad \mathcal{O}_{\traj} (\state_{\obs}, \Omega_\mathrm{yc}, t, \mathcal{V}_\mathrm{{\obs+}}, \mathbf{u}_\mathrm{keep}(t)) = \emptyset , \notag
\end{align}   
where $t_0$ is the current time, $t_\mathrm{end} \in n \, t_\mathrm{m} $ is the time horizon of the maneuver with $n \in \mathbb{N}^+$, $\state_{\ego, t_\mathrm{end}}$ is the final state of the maneuver, and $\mathbf{u}_\mathrm{m}(t)$ is the control sequence for the maneuver trajectory. The predicted obstacle state at $t_\mathrm{end}$ is $\state_{\obs, t_\mathrm{end}}$ and the set $\mathcal{V}_\mathrm{{\obs+}}$ is the spatial extensions of the obstacle enlarged by the safety factor $d_\mathrm{obs,safety}$ for width and length. The occupancy of the obstacle vessel is based on the assumption that the obstacle vessel will keep its speed and course, i.e., the control sequence $\mathbf{u}_\mathrm{keep}(t)$.
This assumption is compliant with the \ac{colregs} collision avoidance rules for the crossing and overtake encounter. In case of the head-on encounter, the predicted trajectory for the obstacle vessel is a conservative prediction since the obstacle vessel would also need to evade to the right to be rule-compliant. Assuming that the obstacle vessel will keep its course and speed leads to the fact that the ego vessel has to turn more to resolve the encounter situation.

\begin{algorithm}[tb]
	\caption{$\mathtt{build\_st}$}
	\label{alg:mt_build}
	\begin{algorithmic}[1]
		\renewcommand{\algorithmicrequire}{\textbf{Input:}}
		\renewcommand{\algorithmicensure}{\textbf{Output:}}
		\REQUIRE candidate action $a_\mathrm{c}$, accelerating actions $\mathcal{A}_\mathrm{acc}$, current state of obstacle vessel $\state_{\obs}$, current state of ego vessel $\state_{\ego}$, maneuver segment time $t_\mathrm{m}$, maneuver horizon $t_\mathrm{max,m}$, control sequence $\mathbf{u}_\mathrm{keep}(t)$\\
		\ENSURE verified part of search tree $\mathcal{G}$\\
		\STATE $t_\mathrm{end} \leftarrow t_\mathrm{m}, \mathcal{G} \leftarrow \{a_\mathrm{c}\}$ \\
		\STATE $\mathbf{u}_\mathrm{c}(t) = \mathtt{a2u}(a_\mathrm{c})$
		\IF{$\mathrm{maneuver\_verified} (\mathbf{u}_\mathrm{c}(t), ...)$}
		\RETURN $\mathcal{G}$ \\
		\ELSE
		\STATE $\mathcal{U}_\mathrm{m} \leftarrow \{\mathbf{u}_\mathrm{c}(t)\}$ 
		\WHILE{$\lnot \mathrm{maneuver\_verified} (\mathbf{u}_\mathrm{m}(t), ...) \; \forall \, \mathbf{u}_\mathrm{m}(t) \in \mathcal{U}_\mathrm{m}$}
		\STATE $\mathcal{U}_\mathrm{m} \leftarrow \emptyset$
		\STATE $\mathcal{G}_\mathrm{temp} \leftarrow \emptyset$
		\IF{$t_\mathrm{end} < t_\mathrm{max,m}$}
		\STATE $t_\mathrm{end} \leftarrow t_\mathrm{end} + t_\mathrm{m}$
		\ELSE
		\RETURN $\mathcal{G} \leftarrow \emptyset$
		\ENDIF
		\FOR{$a^{\prime} \in \mathcal{G}$}
		\IF{$\mathtt{last}(a^{\prime}) = a_\mathrm{c}$}
		\STATE $\mathbf{u}_\mathrm{m}(t) \leftarrow \mathtt{a2u}(a^{\prime}) + \mathtt{a2u}(a_\mathrm{c})$
		\STATE $\mathcal{U}_\mathrm{m} \leftarrow \mathbf{u}_\mathrm{m}(t), \mathcal{G}_\mathrm{temp} \leftarrow [a^{\prime}, a_\mathrm{c}]$
		\FOR{$a_\mathrm{acc} \in \mathcal{A}_\mathrm{acc}$}
		\STATE $\mathbf{u}_\mathrm{m}(t) \leftarrow \mathtt{a2u}(a^{\prime}) + \mathtt{a2u}(a_\mathrm{acc})$
		\STATE $\mathcal{U}_\mathrm{m} \leftarrow \mathbf{u}_\mathrm{m}(t), \mathcal{G}_\mathrm{temp} \leftarrow [a^{\prime}, a_\mathrm{acc}]$
		\ENDFOR
		\ELSE
		\STATE $\mathbf{u}_\mathrm{m}(t) \leftarrow \mathtt{a2u}(a^{\prime}) + \mathtt{a2u}(\mathtt{last}(a^{\prime}))$
		\STATE $\mathcal{U}_\mathrm{m} \leftarrow \mathbf{u}_\mathrm{m}(t), \mathcal{G}_\mathrm{temp} \leftarrow [a^{\prime},\mathtt{last}(a^{\prime})]$
		\ENDIF
		\ENDFOR
		\STATE $\mathcal{G} \leftarrow \mathcal{G}_\mathrm{temp}$
		\ENDWHILE
		\ENDIF
		\RETURN $\mathcal{G}$
	\end{algorithmic}
\end{algorithm}

\begin{figure}
	\centering{
		\resizebox{1.0\columnwidth}{!}{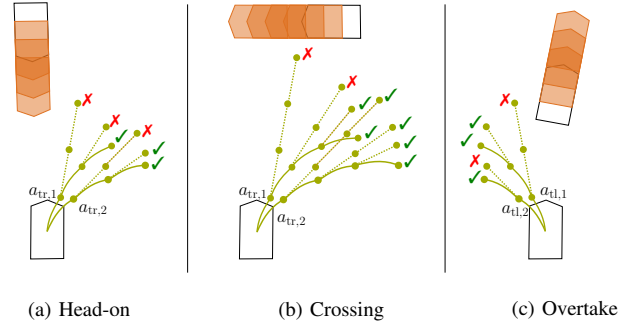}} 
	\caption{Example search trees for the three give-way encounter situations, in which the ego vessel has to give way. The prediction of the obstacle vessel is depicted in orange and the maneuver segment trajectories in green with a dot for the final state. The trajectories based on actions from $\mathcal{A}_\mathrm{acc}$ are displayed as dashed line. Note that we display only one trajectory based on actions from $\mathcal{A}_\mathrm{acc}$ for visualization purposes. The candidate actions initializing the search trees are $a_{d,1}$ and $a_{d,2}$ where $d$ is either $tr$ for turning right and $tl$ for turning left. The mark \textcolor[rgb]{0,0.50196078,0}{\cmark} indicates that the maneuver is verified for the $\mathrm{maneuver\_verified}$ predicate and \textcolor{red}{\xmark} indicates that the maneuver is not rule-compliant.}
	\label{fig:giv_way_verification}
\end{figure}

With the turning direction and the maneuver verification predicate defined in \cref{eq:predicate_maneuver_verified}, we want to determine all actions that lead to verified maneuvers. The generation of maneuvers based on candidate actions is computed by a breadth-first search with rule-compliant pruning. 
The search algorithm is detailed in \cref{alg:mt_build}. Note that to obtain a control sequence for multiple actions, we introduce the function $\mathtt{a2u}$. For a maneuver segment trajectory, the control input corresponding to an action, is held constant for a maneuver segment time $t_\mathrm{m}$ while \cref{eq:yaw_model} is forward simulated.   
We initialize a search tree with a maneuver segment trajectory resulting from the candidate turning action $a_\mathrm{c}$.
A candidate action $a_\mathrm{c}$ ensures that the orientation of the ego vessel changes at least $\Delta_\mathrm{large\_turn}$ within $t_\mathrm{m}$.
Potentially, this first maneuver segment trajectory results already in a verifiable maneuver (cf. \cref{alg:mt_build}, line 2--3). If not, the search tree is extended by (a) a maneuver segment trajectory based on the candidate action $a_\mathrm{c}$ (cf. \cref{alg:mt_build}, line 17--18), and (b) with maneuver segment trajectories for each action $a \in \mathcal{A}_\mathrm{acc}$, which keep the speed or accelerate the ego vessel (cf. \cref{alg:mt_build}, line 19--21). If the action of the maneuver segment trajectory that should be extended (obtained with the function $\mathtt{last}$) does not correspond to $a_\mathrm{c}$, the maneuver is only extended with the previously used action (cf. \cref{alg:mt_build}, line 24--25). This has the effect that the vessel does not switch between different accelerations during the maneuver. The expansion of the search tree is stopped (a) if at least one trajectory sequence is verified for the current search tree depth, i.e., for time horizon $t_\mathrm{end}$, or (b) if the maneuver horizon $t_\mathrm{max,m}$ is reached. Note that $t_\mathrm{max,m}$ follows from the rule specification and is $t_\mathrm{react} + 2 t_\mathrm{maneuver}$. The search tree generation is illustrated in \cref{fig:giv_way_verification} for three give-way encounters.  
Due to the rule-compliant pruning, our search algorithm has the time complexity $\mathcal{O}(n \, N_\mathrm{c} \, N_\mathrm{acc})$ for tree generation where $N_\mathrm{c} \in \mathbb{N}^+$ is the number of candidate actions $a_\mathrm{c}$, and $N_\mathrm{acc} \in \mathbb{N}^+$ is the number of actions in $\mathcal{A}_\mathrm{acc}$.

\paragraph{Actions for encounter maneuvers} \Cref{alg:rule_compliant_actions} summarizes the action verification to achieve rule-compliant maneuvers for rules $R_3$ - $R_6$ given the statechart $\Gamma$ is in an encounter state (i.e., $\exists i \in \{1,...,4\}\colon \mathtt{in}(\rho_i)$). We denote the search tree generation with $\mathtt{build\_st}$ (see \cref{alg:mt_build}) and the detection of actions in the correct turning direction for overtake situations is abbreviated by the function $\mathtt{get\_turning\_act}$. The result of \cref{alg:rule_compliant_actions} is the safe action set $\mathcal{A}_s$ and the verified part of the search tree $\mathcal{G}$. 
\begin{algorithm}[tb]
	\caption{Encounter action verification}
	\label{alg:rule_compliant_actions}
	\begin{algorithmic}[1]
		\renewcommand{\algorithmicrequire}{\textbf{Input:}}
		\renewcommand{\algorithmicensure}{\textbf{Output:}}
		\REQUIRE stand-on action $a_\mathrm{keep}$, turning to right actions $\mathcal{A}_{tr}$, turning to left actions $\mathcal{A}_{tl}$, accelerating actions $\mathcal{A}_\mathrm{acc}$, current state of obstacle vessel $\state_{\obs}$, current state of ego vessel $\state_{\ego}$, encounter predicate $\psi_e$ \\
		\ENSURE set of safe actions $\mathcal{A}_s$, verified part of search tree $\mathcal{G}$\\
		\STATE $\mathcal{A}_s \leftarrow \emptyset, \mathcal{G} \leftarrow \emptyset$ \\
		\IF{$\psi_e = \keep$}
		\STATE $\mathcal{A}_s \leftarrow \{a_\mathrm{keep}\}$ \\
		\ELSE
		\IF{$\psi_e = \headon \lor \psi_e = \crossing$}
		\STATE $\mathcal{A}_\mathrm{temp} \leftarrow \mathcal{A}_{tr}$
		\ELSE
		\STATE $\mathcal{A}_\mathrm{temp} \leftarrow  \mathtt{get\_turning\_act}(\state_{\ego}, \state_{\obs}, \mathcal{A}_{tr}, \mathcal{A}_{tl})$ \\
		\ENDIF
		\FOR{$a \in \mathcal{A}_\mathrm{temp}$}
		\STATE $\mathcal{G}_\mathrm{temp} \leftarrow \mathtt{build\_st}(\state_{\ego}, \state_{\obs}, a, \mathcal{A}_\mathrm{acc}, t_\mathrm{m}, t_\mathrm{max, m})$
		\IF{$\mathcal{G}_\mathrm{temp} \neq \emptyset$}
		\STATE $\mathcal{G}\leftarrow \mathcal{G}_\mathrm{temp}, \mathcal{A}_s \leftarrow a$
		\ENDIF
		\ENDFOR
		\ENDIF
		\RETURN $\mathcal{A}_s, \mathcal{G}$
	\end{algorithmic}
\end{algorithm}

In an encounter situation, in which the ego vessel has to give way, a maneuver of the verified part of the search tree $ \mathcal{G}$ is performed until there is no collision risk with respect to the obstacle vessel. In particular, the actions are conducted for at least the maneuver segment time $t_\mathrm{m}$. At the end of a maneuver segment, the encounter situation is either resolved, or the action selection is constrained to the children of the selected search tree node.
If $\mathcal{G}$ is an empty set, the ego vessel is a stand-on vessel and the only selectable action is $a_\mathrm{keep}$.   

\subsection{Safe-by-design Action Selection} \label{sec:safe_by_design}
We utilize a discrete action space for \ac{rl} since this realizes efficient online safety verification and makes the encounter action verification feasible. In particular, we define an action set $\mathcal{A}$ of $49$ discrete actions. One action represents the emergency action $a_\mathrm{em}$ and the others result from the combination of turning rates and accelerations:
\begin{align}
	\mathcal{A} = &\, \{a_\mathrm{em}, \mathcal{A}_\mathrm{regular}\}  \quad \text{where} \\
	\mathcal{A}_\mathrm{regular} = &\, \{\nacceleration \times \turingrate \, | \nacceleration \in \mathcal{A}_{\nacceleration}, \turingrate \in \mathcal{A}_{\turingrate} \}, \notag
\end{align}
where $\mathcal{A}_{\nacceleration}$ is the finite set describing the allowed normal accelerations and $\mathcal{A}_{\turingrate}$ is the finite set describing the allowed turning rates. 

In the previous sections, we derived the verification of rule-compliant actions. By constraining the \ac{rl} agent to these rule-compliant actions, we ensure by design that only safe actions are executed, and consequently only safe trajectories are performed. \cref{theorem:safe_by_design_action_selection} states the solution to our problem statement in \cref{eq:problemstatement}.

\begin{theorem}\label{theorem:safe_by_design_action_selection}
	Legal safety specified by $\left\langle  \Phi, \leq \right\rangle$ can be ensured through constraining the action space of the \ac{rl} agent to $\mathcal{A}_s(\hat{\rho})$ since all actions in $\mathcal{A}_s(\hat{\rho})$ are specification-compliant actions.
\end{theorem}

\begin{IEEEproof}
	To prove this statement, we derive the safe action set $\mathcal{A}_s$ for all states of the statechart $\Gamma$. \smallskip
	
	\emph{(I) Initial state $\rho_0$:} \enspace Since no rules apply in this state as proven in \cref{theorem:statechart}, any action is compliant with the specification and $\mathcal{A}_s(\rho_0) = \mathcal{A}_\mathrm{regular}$. \smallskip
	
	\emph{(II) Emergency state $\rho_5$:} \enspace We constrain the actions of the \ac{rl} agent to the emergency action $a_\mathrm{em}$ returned by \cref{alg:emergencymaneuvering}, i.e., $\mathcal{A}_s(\rho_5) = a_\mathrm{em}$.\smallskip
	
	\emph{(III) Encounter states $\rho_1 - \rho_4$:} Based on \cref{theorem:statechart} the maneuver predicates for the respective encounter situations must hold in these states to comply with the specification. \Cref{alg:rule_compliant_actions} returns the synthesized rule-compliant maneuvers and respective actions $\mathcal{A}_s(\rho_i)$ where $i \in \{1,...,4\}$. \smallskip
	
	Given $\mathcal{A}_s(\rho)$, we can constrain the action selection of the \ac{rl} agent to $\mathcal{A}_s(\rho)$ with standard action masking \cite{krasowski2023provably} to obtain the safe policy $\pi_s$. Since the safe policy $\pi_s$ only allows rule-compliant actions from $\mathcal{A}_s$, the trajectories $\zeta_{\pi_s}$ are compliant with the legal safety specification $\left\langle  \Phi, \leq \right\rangle$.
\end{IEEEproof}

\section{Reinforcement Learning}\label{ch:reinforcement_learning}
For the task of autonomous vessel navigation on the open sea, we design a simulation environment based on \mbox{CommonOcean} benchmarks \cite{Krasowski2022.CommonOcean} and the yaw-constrained dynamics in (\ref{eq:yaw_model}). The CommonOcean benchmarks contain a planning problem which specifies the goal area and initial state of the ego vessel as well as a scenario which specifies the traffic situation, i.e., for this study the trajectory of the obstacle vessel and the navigational area. At the start of an episode, a CommonOcean benchmark is randomly selected from the training set and the agent is provided with the initial observation. Based on the observation, the agent selects an action from the action set and receives the corresponding reward and next observation of the environment (see \cref{fig:overview_paper}). If the safety verification is activated, the agent can only select from the verified safe action set $\mathcal{A}_s$ as derived in \cref{ch:maneuver_synthesis}. We regard a setting with finite time horizon episodes and terminate the episode in specified situations (see \cref{sec:observation_and_termination}). 
The observation space, termination conditions, action space, action selection constraints, and reward function are detailed in the following paragraphs. 

\subsection{Observation Space and Termination}\label{sec:observation_and_termination}
\begin{figure}
	\centering{
		\resizebox{0.35\textwidth}{!}{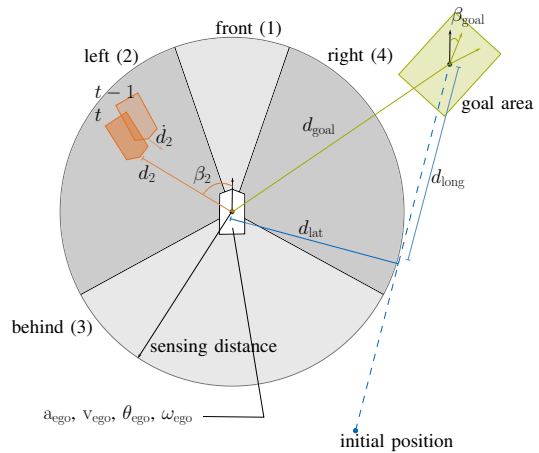}} \vspace{-0.4cm}
	\caption{Illustration of observations with sensing range and four sectors in gray, goal region in green, initial position with direct path to goal region in blue, and obstacle vessel for the previous time step $t-1$ and the current time step $t$ in orange.}
	\label{fig:observation}
\end{figure}

The observation space has $27$ dimensions. We specify four types of observations: ego vessel observations, goal observations, surrounding traffic observations, and termination observations. \Cref{fig:observation} visualizes the ego vessel observations, goal observations, and surrounding traffic observations for the time step $t$. 

The four ego vessel observations are the velocity $\nvelocity_{\ego}$ and orientation $\orientation_{\ego}$ of the ego vessel state $\state_{\ego}$, the acceleration $\nacceleration_{\ego}$, and turning rate $\turingrate_{\ego}$ corresponding to the ego vessel control input. The five continuous goal observations are the Euclidean distance to the goal $d_\mathrm{goal}$, the remaining time steps until the maximal time step of the episode $k_\mathrm{max}$, the orientation difference to the goal orientation range $\beta_\mathrm{goal}$, and the longitudinal $d_\mathrm{long}$ and lateral $d_\mathrm{lat}$ position with respect to the line from the initial state to the center of the goal state. The observations $d_\mathrm{long}$ and $d_\mathrm{lat}$ are relevant since they indicate the deviation of the ego vessel from the optimal path when no other vessels need to be avoided. Additionally, we provide one Boolean goal observation that evaluates to true whenever $\min(|d_\mathrm{lat}|, |d_\mathrm{long}|)$ is larger than the distance $d_\mathrm{hull}$, i.e., the ego vessel is far away from the path between the initial state and goal area.

The surrounding traffic observations are the distance $d_{j}$, angle $\beta_{j}$ and distance rate $\dot{d}_{j}$ for the detected vessel in the sector $j \in \{1, ..., J\}$, where $J$ is the number of sectors. The vessels are only detected if the Euclidean distance to the ego vessel is at most the sensing distance $d_\mathrm{sense}$. For this study, we align the sectors with the sectors specified for the \ac{colregs} collision avoidance rules. Thus, we obtain the four sectors front, left, right, and behind and twelve observation variables, as depicted in \cref{fig:observation}. 

The five termination observations are Boolean observations and indicate if 
\begin{itemize}
	\item the maximal time step was reached $\mathbbm{1}_\mathrm{time} = 1$,
	\item the vessel is outside of the navigational area $\mathbbm{1}_\mathrm{area} = 1$,
	\item the vessel velocity is zero $\mathbbm{1}_\mathrm{stopped} = 1$,
	\item the vessel collided $\mathbbm{1}_\mathrm{collision} = 1$,
	\item the vessel reached the goal area $\mathbbm{1}_\mathrm{goal} = 1$.
\end{itemize}
We terminate the episode when the ego vessel stopped, as reverse driving is not meaningful on the open sea and the termination leads to the agent being reset to a much more meaningful initial state of another CommonOcean benchmark. The termination conditions follow directly form the termination observations, as we terminate the episode if one of these observations is present. 

\subsection{Reward}
The reward is designed such that the vessel is reinforced in goal reaching behavior and penalized for unsafe or inefficient behavior. In particular, we design a reward function based on sparse and dense components. 
The sparse rewards are related to termination conditions and using the emergency planner:
\begin{align*}
	r_\mathrm{sparse} = \, &c_\mathrm{time} \mathbbm{1}_\mathrm{time} + c_\mathrm{area} \mathbbm{1}_\mathrm{area} + c_\mathrm{goal} \mathbbm{1}_\mathrm{goal} \, + \\
	&c_\mathrm{stopped} \mathbbm{1}_\mathrm{stopped} + c_\mathrm{collision} \mathbbm{1}_\mathrm{collision} \, + \\
	&c_\mathrm{emergency} \mathbbm{1}_\mathrm{emergency},
\end{align*}
where $c_\mathrm{i}$ indicate the reward coefficients, which are all negative except for $c_\mathrm{goal}$.

Additionally, we define four types of dense rewards for \ac{colregs} compliance, advancing to the goal, keeping the velocity, and deviation from the path between initial state and goal. To incentivise behavior that is compliant with the collision avoidance rules specified in the \ac{colregs}, we utilize a reward component specified in \cite[Eq. (26)]{Meyer2020}:
\begin{equation*}
	r_\mathrm{colregs} = - \frac{\alpha}{1 + \exp(\gamma_\mathrm{\phi,dyn} |\phi|)}  \exp((\zeta_v v_\mathrm{\obs,\phi} - \zeta_\mathrm{\obs,d}) d_{\obs}).
\end{equation*}
The angle $\phi \in [-\pi, \pi]$ specifies the relative angle between the ego orientation and the orientation toward the obstacle vessel, $v_\mathrm{obs,\phi}$ specifies the velocity component of the obstacle vessel velocity in the radial direction from the ego vessel to the obstacle vessel, and $d_{\obs}$ is the distance observed to the obstacle vessel, i.e., the respective $d_{j}$. The parameters $\alpha, \gamma_\mathrm{\phi,dyn}, \zeta_v$, and $\zeta_\mathrm{obs,d}$ are set to the same values as defined in \cite{Meyer2020}.

Further, we define a reward component that supports the agent in learning how to reach the goal by providing a reward that is proportional to the advance or retreat from the goal since the previous time step:
\begin{align*}
	r_\mathrm{goal} = & c_\mathrm{reach} \left( \|\position_{\ego, t} - \position_\mathrm{goal} \|_{2} - \|\position_{\ego, t-1} - \position_\mathrm{goal} \|_2 \right).
\end{align*}
The center position of the goal area is $\position_\mathrm{goal}$, and $\position_{\ego, t}$ is the current ego position, $\position_{\ego, t-1}$ is the ego vessel position at the previous time step, and $c_\mathrm{reach}$ is a scaling coefficient.  

On the open sea, vessels typically navigate in a narrow speed range. To enforce this also for the \ac{rl} agent, the reward component $r_\mathrm{velocity}$ provides a penalty proportional to the deviation from the desired speed range:
\begin{equation*}
	r_\mathrm{velocity} = \begin{cases}
	    c_v (\nvelocity_{\ego} - \nvelocity_\mathrm{high}) & \text{if } \nvelocity_{\ego} > \nvelocity_\mathrm{high}\\
	    c_v (\nvelocity_\mathrm{low} - \nvelocity_{\ego}) & \text{if }  \nvelocity_{\ego} < \nvelocity_\mathrm{low}\\
	    0& \text{otherwise}.
	\end{cases}
\end{equation*}
The parameters $\nvelocity_\mathrm{low}$ and $\nvelocity_\mathrm{high}$ define the speed range bounds, and $c_v$ is the reward coefficient.

The last reward component informs the agent about its deviation from the direct path between the initial state and the goal area: 
\begin{equation*}
	r_\mathrm{deviate} = c_\mathrm{deviate} \min(|d_\mathrm{lat}|, d_\mathrm{hull}),
\end{equation*}
where the coefficient $c_\mathrm{deviate}$ scales the penalty proportional to the absolute lateral deviation $|d_\mathrm{lat}|$, and $c_\mathrm{deviate} d_\mathrm{hull}$ is the maximum of the reward component $r_\mathrm{deviate}$. Finally, the reward function is given by the sum of all components:
\begin{align}
	r = r_\mathrm{sparse} + r_\mathrm{colregs} + r_\mathrm{goal} + r_\mathrm{velocity} + r_\mathrm{deviate}. \label{eq:reward}
\end{align}

\section{Numerical Experiments}
\label{ch:experiments}

\begin{table}[tb]
	\vspace{0.2cm}
	\caption{Experimental parameters}
	\renewcommand{\arraystretch}{1.3}
	\centering
	\begin{tabular}{lrlr}
		\toprule
		\textbf{Parameter} & \textbf{Value} & \textbf{Parameter} & \textbf{Value} \\ \midrule
		\multicolumn{4}{l}{\emph{Safety verification}} \\ \midrule
		$\Delta_\mathrm{ahead}$& \SI{45}{\deg} & $\Delta_\mathrm{stern}$& \SI{20}{\deg} \\
		$\nvelocity_\mathrm{pm,max}$ & \SI{10}{\meter \per \second}& $\nacceleration_\mathrm{pm,max}$ & \SI{0.045}{\meter \per \second \squared}\\
		$d_{\mathrm{resolved}}$ & 2 $l_{\ego}$ &$\nacceleration_\mathrm{stern}$ & 0.2 $\nacceleration_\mathrm{max}$ \\ 
		$d_\mathrm{obs,safety}$ & 2 $l_{\obs}$ & $d_\mathrm{min, ahead}$& 3 $l_{\obs}$\\
		$\Delta_{\mathrm{head\text{-}on}}$& \SI{5}{\deg}& $t_\mathrm{horizon}^\mathrm{check}$& \SI{420}{\second}\\
		$\Delta_\mathrm{no\_turn}$& \SI{10}{\deg}&  $t_\mathrm{maneuver}$& \SI{70}{\second}\\
		$\Delta_\mathrm{large\_turn}$& \SI{20}{\deg}& $t_\mathrm{react}$ &\SI{60}{\second} \\ 
		$t_\mathrm{pred}$& \SI{180}{\second} & $t_\mathrm{m}$ & \SI{40}{\second} \\ 
		$t_\mathrm{max,m}$& \SI{200}{\second} & & \\ \midrule
		\multicolumn{4}{l}{\emph{Ego vessel}} \\ \midrule
		$l_{\ego}$ & \SI{175}{\meter}& $\turingrate_{\mathrm{max}}$& \SI{0.03}{\radian \per \second} \\ 
		$\nacceleration_\mathrm{max}$ &  \SI{0.24}{\meter \per \second \squared}& $v_{\mathrm{max}}$& \SI{9.5}{\meter \per \second} \\ \midrule
		\multicolumn{4}{l}{\emph{Reinforcement learning}} \\ \midrule
		$\nvelocity_{\mathrm{low}}$ & \SI{2.5}{\meter \per \second}& $\nvelocity_{\mathrm{high}}$& \SI{8}{\meter \per \second} \\ 
		$c_{\mathrm{time}}$ & \SI{-25}{}& $c_{\mathrm{area}}$ & \SI{-5}{}\\ 
		$c_{\mathrm{goal}}$ & \SI{50}{}& $c_{\mathrm{stopped}}$ & \SI{-40}{} \\
		$c_{\mathrm{collision}}$ & \SI{-50}{}& $c_{\mathrm{emergency}}$ & \SI{-0.5}{}\\
		$c_{\mathrm{reach}}$ & \SI{1.5}{}& $c_{{v}}$ & \SI{-2}{} \\
		$c_{\mathrm{deviate}}$ & \SI{-0.001}{}&  $d_{\mathrm{sense}}$ & \SI{8000}{\meter} \\
		$d_{\mathrm{hull}}$ & \SI{2000}{\meter}&  $J$ & \SI{4}{} \\
		\multicolumn{4}{l}{$\mathcal{A}_{\nacceleration} = \{-0.048, -0.032, -0.016, 0, 0.016, 0.032, 0.048\} \si{\meter \per \second \squared}$} \\ 
		\multicolumn{4}{l}{$\mathcal{A}_{\turingrate} = \{-0.018, -0.012, -0.06, 0, 0.06, 0.012, 0.018\} \si{\radian \per \second}$} \\ \bottomrule
	\end{tabular}
	\label{tab:parameters}
\end{table} 

Critical encounter situations are rare in maritime traffic data. Thus, this data is not well suited for training \ac{rl} agents that should learn how to handle encounter situations. Therefore, we construct random CommonOcean benchmarks \cite{Krasowski2022.CommonOcean} that represent critical encounters as a foundation of our simulation environment. In particular, we initialize the ego vessel and the other vessel approximately $\SI{2000}{\meter}$ - $\SI{3500}{\meter}$ away from their closest encounter position. The initial velocity range for both vessels is $[\SI{3}{\meter \per \second}, \SI{7}{\meter \per \second}]$. For the obstacle vessel, we generate a trajectory that is close to constant velocity and speed, and disturb the initial orientation and velocity with values sampled uniformly from $[\SI{-0.05}{\radian}, \SI{0.05}{\radian}]$ and $[\SI{-0.1}{\meter \per \second}, \SI{0.1}{\meter \per \second}]$, respectively, to make the trajectory more realistic. The goal area is approximately $\SI{4500}{\meter}$ away from the initial position of the ego vessel. The goal area is $\SI{400}{\meter}$ long and $\SI{60}{\meter}$ wide. The time horizon for the scenario is $k_\mathrm{max} = 170$ time steps where the time step size is $\Delta t = \SI{10}{\second}$. In total, we constructed $2000$ CommonOcean benchmarks \cite{Krasowski2022.CommonOcean} and randomly split them in a $\SI{70}{\percent}$ training and $\SI{30}{\percent}$ testing set. The model of the ego vessel is the yaw-constrained model in \cref{eq:yaw_model} and we use the parameters of a container vessel\footnote{The container vessel is the vessel type 1 from \href{https://commonocean.cps.cit.tum.de/commonocean-models}{commonocean.cps.cit.tum.de/commonocean-models}.}. We reduce the maximum velocity specified in the vessel parameters to $\SI{9.5}{\meter \per \second}$ to better match a realistic velocity range for open sea maneuvering.  

Next to the simulation environment, we need to specify values for the parameters of the safety verification approach, ego vessel, and reinforcement learning. \Cref{tab:parameters} summarizes the parameters. Note that the emergency controller can use the full control input space specified for the ego vessel through the intervals $[-\nacceleration_\mathrm{max}, \nacceleration_\mathrm{max}]$ and $[-\turingrate_{\mathrm{max}}, \turingrate_{\mathrm{max}}]$. For normal operation, we reduce the control input limits to a more reasonable range for open sea maneuvering. This is reflected by the sets of allowable accelerations $\mathcal{A}_{\nacceleration}$ and turning rates $\mathcal{A}_{\turingrate}$ (see \cref{tab:parameters}).
As model-free \ac{rl} algorithm, we used proximal policy optimization (PPO) \cite{Schulman.2017}. Our implementation is based on stable-baselines3 \cite{stable-baselines3} and the action masking implementation in \cite{krasowski2023provably}. The agent networks are multi-layer perceptron networks with two layers and 64 neurons in each layer.

\subsection{Evaluation concept}
To comprehensively evaluate our approach, we introduce two benchmark agents next to our provably safe agent and compare different deployment setups. We train all three agents in our simulation environment, which is based on the training data of critical CommonOcean benchmarks \cite{Krasowski2022.CommonOcean}. The trained agents are:
\begin{enumerate}
	\item the \emph{baseline} agent with the reward function $r =  r_\mathrm{sparse} + r_\mathrm{goal} + r_\mathrm{velocity} + r_\mathrm{deviate}$, i.e., $r_\mathrm{colregs} = 0$ in \cref{eq:reward}, and no safety verification,
	\item the \emph{rule-reward} agent, which is informed by the \ac{colregs} reward $r_\mathrm{colregs}$, i.e., reward function \cref{eq:reward}, and
	\item the \emph{safe} agent with safety verification and reward function \cref{eq:reward}.
\end{enumerate}
The baseline agent represents a straightforward \ac{rl} implementation for which the agent is informed about unsafe actions only sparsely with a collision penalty. The rule-reward agent models the state-of-the-art for traffic-rule-informed open-sea vessel navigation \cite{Heiberg2022, Chun2021, Meyer2020, Xu2022}, because the reward function includes a \ac{colregs} reward $r_\mathrm{colregs}$. 
For each agent type, we use ten random seeds and train an agent per seed for three million environment steps.

We evaluate the deployment performance of the trained agents on the testing set of the handcrafted critical scenarios and on scenarios from recorded traffic data\footnote{All scenarios are publicly on the \href{https://commonocean.cps.cit.tum.de/}{CommonOcean website} with ids ZAM\_AAA-1\_20240121\_T-[0,...,1999].}. For the rule-reward and baseline agent, we investigate performance without, i.e., as trained, and with safety verification enabled. Including the safety verification after training allows us to evaluate if guaranteeing traffic rule compliance after training is sufficient. Note that the action space of the two benchmark agents is $\mathcal{A}_\mathrm{regular}$, except for deployment with safety verification.

We consider critical scenarios from recorded traffic data to examine the generalization of the agents to real-world situations. To this end, we use marine traffic data from three large open-sea areas off the US coast from \cite{Krasowski.2021} and extract critical encounters. In particular, we only use scenarios where the distance between two vessels drops to $\SI{5000}{\meter}$ or lower. Further, we ensure that the paths of both vessels cross each other. Then, we replace one vessel by an ego vessel to generate the initial state and goal area. The initial state is part of the recorded trajectory and is selected about $\SI{2000}{\meter}$ before the closest encounter. The position of the goal area is also part of the recorded trajectory and is about $\SI{2000}{\meter}$ after the closest encounter. We use the same shape for the goal area as in our handcrafted scenarios. In total, we identify 49 critical scenarios in the three large open-sea areas off the US coast from traffic data of January 2019 (about 30 GB of raw Automatic Identification System (AIS) data).

We evaluate our agents based on the goal-reaching rate, reward, episode lengths, collisions, emergency controller usage and rule violations. Rule violations reflect how often per episode the regular collision avoidance rules are violated. For that, we count:
\begin{itemize}
	\item every time step of violating the stand-on vessel position results;
	\item every crossing, overtaking and head-on encounter for which no proper collision avoidance maneuver is taken.
\end{itemize}

\subsection{Results}

\begin{figure}
	\centering
	\includegraphics[width=0.48\columnwidth]{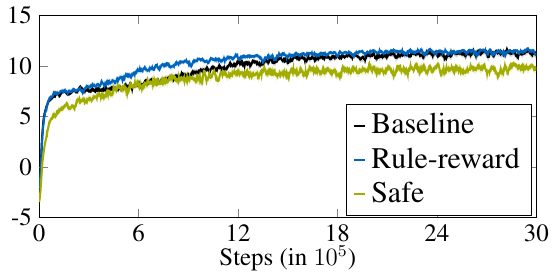}%
	\hfil
	\includegraphics[width=0.485\columnwidth]{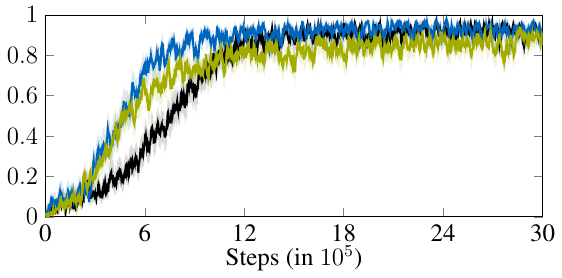}\\ \vspace{-0.2cm}
	{\scriptsize (a) Average reward  \hspace{2.2cm} (b) Goal-reaching rate \hspace{0.4cm}}\\ \vspace{0.2cm}
	\includegraphics[width=0.5\columnwidth]{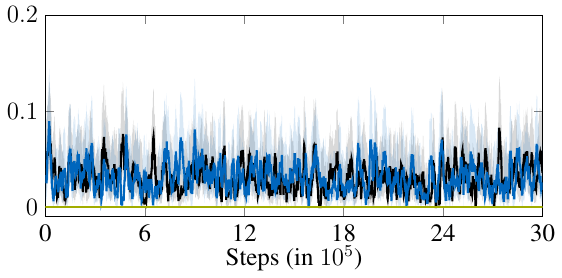}%
	\hfil
	\includegraphics[width=0.485\columnwidth]{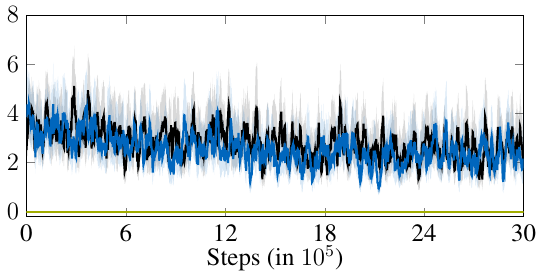}\\ \vspace{-0.2cm}
	{\scriptsize \hspace{0.15cm} (c) Collision rate \hspace{2.35cm} (d) Rule violations }
		\caption{Mean and bootstrapped $\SI{95}{\percent}$ confidence interval for training curves for baseline, rule-reward, and safe agents averaged over ten random seeds.}
		\vspace{-0.5cm}
		\label{fig:results_training}
\end{figure}

\paragraph{Training evaluation} \Cref{fig:results_training} shows the training curves for the three agent types. The average reward curves show similar convergence across agent types, although the baseline and rule-reward agents achieve slightly higher rewards after three million training steps. Note that for the displayed reward curves, the emergency penalty and \ac{colregs} reward term $r_\mathrm{colregs}$ are subtracted for comparability. The goal-reaching rate curves mirror the reward curves and the agents reach goals in about $\SI{90}{\percent}$ of all scenarios at the end of the training. We observe that the agent types without safety verification reach the goal slightly more often.

Importantly, there are no collisions and rule violations for the safe agent (see \cref{fig:results_training}c and \cref{fig:results_training}d). For the baseline and rule-reward agent, the collision rate is relatively stable around $\SI{5}{\percent}$ during the full training time. 
Rule violations for the baseline and rule-reward agent slightly decrease but never reach zero. This suggests that complying with the \ac{colregs} effectively achieves collision avoidance.

\paragraph{Deployment evaluation} The results averaged over ten random seeds for each agent type are summarized in \cref{tab:deploy_results}. For the handcrafted scenarios, the rule-reward agents reach the goal for $\SI{90.7}{\percent}$ of the scenarios. This is about $\SI{5}{\percent}$ higher than for the baseline and safe agent. Yet, only the safe agent achieves zero collisions and no rule violations. The rule-reward agent collides and violates the rules fewer times than the baseline agent. If the safety verification is enabled for the baseline and rule-reward agent, the goal-reaching rate drops significantly by approximately $\SI{40}{\percent}$. Additionally, for the safe agent, the emergency controller intervenes on average in $\SI{6}{\percent}$  of the time steps in an episode, whereas for the rule-reward and baseline agents with activated safety verification, the emergency controller is needed in approximately $\SI{10}{\percent}$ of the time steps in an episode.

\Cref{tab:deploy_results} displays the testing results on the 49 critical recorded traffic scenarios for the different agent types. The rule-reward agent reaches the goal most often and exhibits the lowest average episode length. Interestingly, the goal-reaching rate for the baseline and rule-reward agent drops only by about $\SI{5}{\percent}$ when activating our safety verification approach. The collision rate and rule violation rate are smaller than for the handcrafted scenarios. With activated safety verification, we observe no collisions and no rule violations. 
Note that the differences in the reported means for goal-reaching rate and emergency steps between the agents with activated safety verification are statistically insignificant\footnote{The p-values for paired t-tests between the safe agent and the rule-reward and baseline agents for the goal-reaching rate are 0.248 and 0.951, respectively. The p-values for agents with respect to emergency steps are 0.211 and 0.381.}.

\begin{table}[tb]
	\vspace{0.2cm}
	\caption{Testing results on 600 handcrafted and 49 recorded scenarios}
	\renewcommand{\arraystretch}{1.2}
	\centering
	\begin{tabular}{@{\extracolsep{2pt}}lcccccc@{}}
		\toprule
		\multicolumn{2}{c}{\textbf{Setup}} & \multicolumn{2}{c}{\textbf{Efficiency}} &  \multicolumn{3}{c}{\textbf{Safety}} \\ \cmidrule{1-2} \cmidrule{3-4} \cmidrule{5-7}
		{Agent} & \makecell[c]{{Verify}} & \makecell[c]{Goal- \\ reach} & \makecell[c]{Ep. \\ length}  & {Collided} & \makecell[c]{Rules \\ violated} & 
		\makecell[c]{Emerg. \\ steps}\\ \midrule
		\multicolumn{7}{l}{\emph{Handcrafted testing scenarios}} \\ \midrule
		Base & \xmark &\SI{86.8}{\percent}& \SI{566}{\second} &\SI{3.13}{\percent} & \SI{2.65}{} & -- \\ 
		RR & \xmark & $\mathbf{90.7} \,$\si{\percent} & $\mathbf{544} \,$\si{\second} &\SI{2.85}{\percent} & \SI{2.24}{} & --\\ 
		Base & \cmark &\SI{44.0}{\percent}& \SI{678}{\second} &  $\mathbf{0.0} \,$\si{\percent} & $\mathbf{0}$ & \SI{10.06}{\percent}\\ 
		RR & \cmark & \SI{47.6}{\percent}& \SI{702}{\second} &$\mathbf{0.0} \,$\si{\percent}  & $\mathbf{0}$ & \SI{9.96}{\percent}\\ 
		Safe &  \cmark &\SI{86.3}{\percent}& \SI{647}{\second} & $\mathbf{0.0} \,$\si{\percent}  & $\mathbf{0}$ & $\mathbf{6.18} \,$\si{\percent} \\  \midrule  
		\multicolumn{7}{l}{\emph{Recorded maritime traffic scenarios}} \\ \midrule
		Base & \xmark &\SI{83.1}{\percent}& \SI{563}{\second} &\SI{0.41}{\percent}  & \SI{0.75}{} &-- \\ 
		RR & \xmark & $\mathbf{84.7} \,$\si{\percent} & $\mathbf{550} \,$\si{\second} &\SI{0.41}{\percent}  & \SI{0.82}{} &--\\ 
		Base & \cmark & \SI{78.3}{\percent}& \SI{591}{\second} &$\mathbf{0.0} \,$\si{\percent}  & $\mathbf{0}$ & \SI{2.35}{\percent} \\ 
		RR & \cmark & \SI{82.4}{\percent}& \SI{565}{\second} &$\mathbf{0.0} \,$\si{\percent}  & $\mathbf{0}$ & $\mathbf{1.84} \,$\si{\percent} \\ 
		Safe & \cmark &\SI{78.2}{\percent}& \SI{630}{\second} &$\mathbf{0.0} \,$\si{\percent}  & $\mathbf{0}$ & \SI{2.98}{\percent} \\  \bottomrule 
	\end{tabular}
	\begin{center}
		{\footnotesize \textit{Note: The rule-reward and baseline agents are abbreviated with RR and base. Ep. length is the average episode time horizon. Emerg. steps denote the percentage of steps for which the emergency controller intervened.}}
	\end{center}
	\vspace{-0.5cm}
	\label{tab:deploy_results}
\end{table}

\subsection{Discussion}

\paragraph{Safety in handcrafted scenarios} The safety verification ensures that the encounter traffic rules are never violated and we empirically observe that no collisions occur. However, this results in a lower goal-reaching rate than for the soft-constrained rule-reward agent. One reason for this observation might be that with safety verification, the task is more difficult to solve since the agent is often constrained to avoidance maneuvers before it can maneuver freely again. Thus, the safe agent can explore less freely compared to the baseline and rule-reward agents. The drop in the goal-reaching rate when the safety verification is enabled after training is likely due to the distribution shift, as the baseline and rule-reward agents are probable led to states that they explored less frequently or not at all during training.

\paragraph{Safety on recorded scenarios} In contrast, testing the rule-reward and baseline agent with safety verification on the scenarios from recorded traffic data does not lead to such a significant drop. At the same time, the agent setups without safety verification exhibit fewer rule violations and fewer collisions on the recorded maritime traffic scenarios. Both observations indicate that the scenarios based on recorded data are less critical than the handcrafted situations and, thus, easier to solve for the agents that were not constrained to rule-compliant actions during training. Generally, the agents generalize well to the scenarios based on recorded data. Since identifying critical situations in recorded maritime traffic data is computation-heavy and critical situations are very rare, this small gap between realistic recorded and randomly handcrafted situations is compensated by being able to create many scenarios: The 49 critical situations resulted from one month of maritime traffic data at the coast of the US, whereas the 2000 handcrafted critical situations were generated in a matter of minutes. Yet, recorded scenarios are not fully representing the variety of the real world. Thus, future work should investigate if our safe agent also performs well on a real-world test bed.

\paragraph{Requirements for multi-vessel traffic situations} 
Real-world traffic situations can include more than two vessel on a collision course. Our formalized traffic rules can be evaluated for these more complex traffic situations as demonstrated in \cite{Krasowski.2021}. Yet, the current version of the \ac{colregs} does not provide a clear collision avoidance specification if more than two vessels are involved. Thus, a formal verification cannot be developed due to the lack of a clear specification. Future work should investigate extensions of the \ac{colregs} to fill this specification gap and consequently realize provably rule-compliant motion planning in multi-vessel traffic situations.

\paragraph{Action space choice} The discrete action space makes it possible to efficiently identify rule-compliant actions. However, a continuous action space would allow the agent to explore all possible actions. This significantly increases the challenge of identifying safe actions, because there are infinitely many individual continuous actions in a continuous action space. Yet, one approach to investigate in future work could be obtaining rule-compliant state sets as proposed in \cite{Liu2023} and correcting actions proposed by the agent to safe actions, e.g., with action projection as in \cite{Kochdumper2023.safeRLReachabilityAnalysis}.  

\paragraph{Satisfiablity of rules} The parametrization of the temporal logic rules eases re-adjusting to regulation changes. Yet, these parameters must be manually tuned to ensure that the temporal logic rules are satisfiable. For example, it is important that the detection of an encounter situation happens early enough so that no emergency situation is detected during a give-way maneuver. For instance, theorem provers could help to verify that the chosen rule parameters guarantee that the rules are satisfiable. However, formulating this proof is challenging due to the continuous state and action space, and subject to future work. 

\section{Conclusion}
\label{ch:conclusions}
We are the first to propose a provably safe \ac{rl} approach for autonomous power-driven vessels on the open sea that achieves provable compliance with traffic rules formalized with temporal logic. For that, we introduced an online verification approach based on our formalized rules identifying the set of safe actions. Our formal emergency detection and emergency controller achieves collision avoidance for the regarded traffic situations even if other vessels do not comply with traffic rules. In critical maritime traffic situations, our safe \ac{rl} agent achieves rule compliance, in contrast to state-of-the art agents that are informed about safety only through the reward. At the same time, all agents achieve a satisfactory goal-reaching performance on critical traffic situations. Our evaluation on recorded traffic situations shows that our safe \ac{rl} agent generalizes beyond the distribution of training data. This study is a first step toward learning-based motion planning systems complying with traffic rules for autonomous vessel navigation.

%

\section*{Acknowledgment}

The authors gratefully acknowledge the partial financial support of this work by the research training group ConVeY funded by the German Research Foundation under grant GRK 2428 and by the project TRAITS funded by the German Federal Ministry of Education and Research.




\bibliographystyle{IEEEtran}
\bibliography{literature}


\begin{appendices}

\section*{Appendix}

\subsection{Predicates specified \cite{Krasowski.2021}}\label{appendix:predicates}

In \cref{tab:predicates}, we briefly recapitulate the predicates specified in \cite{Krasowski.2021}. We refer the interested reader to our previous work~\cite{Krasowski.2021} for detailed explanations. Subsequently, the necessary notation that was not yet introduced in this article is introduced and the re-parametrization of the predicate $\collisionpossible$ is explained. 

The trajectory of vessel $i$ consists of states at discrete time steps and is denoted as $\mathcal{T}_i$. The velocity vector based on the state of the vessel is $\mathbf{v}_i = \mathtt{proj}_{\nvelocity}(\state_i) \; \mathtt{unit\_v}(\state_i)$. We define a clock $\mathtt{cl}(\mathcal{T}_i, \state_i)$ that starts at the initial time step of a trajectory and returns the elapsed time for a state $\state_i$. Further, we require a function $\; \mathtt{state}(\mathcal{T}_i, t_k) \;$ which returns the state of a trajectory at time $t_k$. The modulo operator $\mathrm{mod}(a,b)$ returns the remainder of $a / b$ for $a,b \in \mathbb{R}$ using floored division. The function $\mathtt{t}_s$ returns the time for a predicate trace where the respective predicates changed last from false to true. The collision cone $CC'$ is based on the velocity obstacle concept \cite{Fiorini.1998} and the construction is detailed in \cite[Fig.~1]{Krasowski.2021}.

For this work, we made two re-parametrizations of the predicate $\collisionpossible$, which determines if two vessels $l$ and $m$ are on a collision course and, thus, could collide within the time $t_\mathrm{horizon}$.
First, we also want to detect a collision course if the vessels would pass each other with insufficient distance. Thus, we use $r_m = 3 \, l_m$ for the collision cone $CC'$ instead of $r_m = l_m$ in \cite[Fig.~1]{Krasowski.2021}.
This results in detecting a collision possibility if the vessels would not keep a safe distance of at least two lengths of the vessel $m$. 
Second, we evaluate the set of vessel velocities $\mathcal{V}_l$ with respect to their collision possibility instead of only the current velocity $\mathbf{v}_l$. In particular, we check the collision possibility for 
\begin{align*}
	\mathcal{V}_l = \{\lambda \, \mathtt{unit\_v}(\state_{l})| \lambda \in [\mathtt{proj}_{\nvelocity} (\state_{l}) - \nvelocity_{\epsilon}, \mathtt{proj}_{\nvelocity} (\state_{l}) + \nvelocity_{\epsilon}] \}.
\end{align*}
We set the velocity difference $\nvelocity_\mathrm{\epsilon}$ to \SI{1}{\meter \per \second} for our numerical evaluations.\\ 

\subsection{Proof of \cref{lemma:mutal_exclusive}}\label{appendix:lemma} 
\begin{IEEEproof}
	To prove that only one predicate of $\keep$, $\crossing$, $\headon$, and $\overtake$ can evaluate to true, we show for each combination that the conjunction is false when evaluated for two vessels $l$ and $m$.
	For the combination of $\crossing$ and $\headon$, it directly follows that the predicates cannot be true at the same time from the relative position detected by the respective sector predicates.\smallskip
	
	\emph{(I) $\crossing \land \headon$}: 
	\begin{align*}
		&\crossing(\state_l, \state_m,\cdot) \land \headon(\state_l, \state_m,\cdot)  \\
		=&\left(\mathrm{in\_right\_sector}(\state_l, \state_m) \land ... \right)\, \land \\
		&\left(\mathrm{in\_front\_sector}(\state_l, \state_m) \land ...  \right) \\
		=& \,  \bot 
	\end{align*}
	
	For the combination of $\crossing$ and $\overtake$, let us assume that crossing predicate is true. Then, the vessel $m$ is oriented towards left and in the right sector of vessel $l$ (see Fig.~3 and Fig.~4 in \cite{Krasowski.2021}). Thus, it is geometrically impossible for vessel $l$ to be in the behind sector of vessel $m$ and $\overtake$ cannot be true.\smallskip
	
	\emph{(II) $\crossing \land \overtake$}: 
	\begin{align*}
		&\crossing(\state_l, \state_m,\cdot) \land \overtake(\state_l, \state_m,\cdot) \\
		=&\big(\mathrm{in\_right\_sector}(\state_l, \state_m) \land \\
		& \mathrm{orientation\_towards\_left} (\state_l, \state_m, \Delta_{\mathrm{head\text{-}on}}) \land ... \big) \, \land \\
		&\left(\mathrm{in\_behind\_sector}(\state_m, \state_l) \land ...  \right) \\
		=& \, \bot  
	\end{align*}
	
	The predicates $\headon$ and $\overtake$ cannot be true simultaneously as the relative positions and orientations contradict each other similar to case (II). In particular, if the vessel $m$ is in the front sector of vessel $l$ and their relative orientation is in $[\pi - \Delta_{\mathrm{head\text{-}on}}, \pi + \Delta_{\mathrm{head\text{-}on}}]$, then vessel $l$ cannot be in the behind sector of vessel $m$.\smallskip
	
	\emph{(III) $\headon \land \overtake$}: 
	\begin{align*}
		&\headon(\state_l, \state_m,\cdot) \land \overtake(\state_l, \state_m,\cdot) \\
		=&\big(\mathrm{in\_front\_sector}(\state_l, \state_m) \land \\
		&\lnot \mathrm{orientation\_delta}(\state_l, \state_m, \Delta_{\mathrm{head\text{-}on}}, \pi) \land ... \big) \, \land\\
		&\left(\mathrm{in\_behind\_sector}(\state_m, \state_l) \land ...  \right) \\
		=& \, \bot 
	\end{align*}
	
	The predicate $\keep$ is a disjunction of two cases in which the vessel has to keep its course and speed. Thus, we have to show that for both statements of the disjunction that they evaluate to false. The explanation for the equation steps are marked with small letters in round brackets, e.g., (a), and follow after the respective equations. \smallskip
		
	\emph{(IV) $\overtake \land \keep$}: 
	\begin{align*}
		&\overtake(\state_l, \state_m,\cdot) \land \keep(\state_l, \state_m,\cdot) \\
		\overset{\mathrm{(a)}}{=}& \left(\overtake(\state_l, \state_m,\cdot)  \land \left(\mathrm{in\_left\_sector}(\state_l, \state_m) \land ...\right)  \right)  \lor \\
		& \left(\overtake(\state_l, \state_m,\cdot)  \land \overtake(\state_m, \state_l,\cdot) \right) \\
		\overset{\mathrm{(b)}}{=}& \Big(\big(\mathrm{in\_behind\_sector}(\state_l, \state_m) \land ... \big) \, \land \\
		&\big(\mathrm{in\_left\_sector}(\state_l, \state_m) \land ...\big)  \Big) \, \lor  \\ 
		& \left(\overtake(\state_l, \state_m,\cdot)  \land \overtake(\state_m, \state_l,\cdot) \right)\\
		\overset{\mathrm{(c)}}{=}& \, \bot   \lor  \bot   \\
		=& \, \bot 
	\end{align*}
	(a) We distribute the disjunction in $\keep$ over the conjunction with $\overtake$.
	
	\noindent (b) We insert the relevant parts of the predicates (see \cref{tab:predicates}).
	
	\noindent (c) For the first part of the disjunction, the vessels cannot be simultaneously in two sectors as in case (I). For the second part of the disjunction, the two overtake predicates cannot be true at the same time, as both vessels cannot overtake each other at the same time.\smallskip
	
	\emph{(V) $\crossing \land \keep$}: 
	\begin{align*}
		&\crossing(\state_l, \state_m,\cdot) \land \keep(\state_l, \state_m,\cdot) \\
				\overset{\mathrm{(a)}}{=}& \left(\crossing(\state_l, \state_m,\cdot)  \land \left(\mathrm{in\_left\_sector}(\state_l, \state_m) \land ...\right)  \right)  \lor \\
		& \left(\crossing(\state_l, \state_m,\cdot)  \land \overtake(\state_m, \state_l,\cdot) \right) \\
		\overset{\mathrm{(b)}}{=}& \Big(\big(\mathrm{in\_right\_sector}(\state_l, \state_m) \land ... \big) \, \land \\
		&\big(\mathrm{in\_left\_sector}(\state_l, \state_m) \land ...\big)  \Big) \, \lor \\
		& \Big( \big( \mathrm{in\_right\_sector} (\state_l, \state_m) \, \land ... \big)  \, \land \\
		& \big( \mathrm{in\_behind\_sector}(\state_l, \state_m) \land ... \big)\Big)\\
		\overset{\mathrm{(c)}}{=}&  \, \bot   \lor  \bot   \\
		=& \, \bot  
	\end{align*}
	(a) We distribute the disjunction in $\keep$ over the conjunction with $\crossing$.
	
	\noindent (b) We insert the relevant parts of the predicates (see \cref{tab:predicates}).
	
	\noindent (c) For both parts of the disjunction, the vessels cannot be simultaneously in two sectors as in case (I).\smallskip
	
	\emph{(VI) $\headon \land \keep$}: 
	\begin{align*}
		&\headon(\state_l, \state_m,\cdot) \land \keep(\state_l, \state_m,\cdot) \\
		\overset{\mathrm{(a)}}{=}& \left(\headon(\state_l, \state_m,\cdot)  \land \left(\mathrm{in\_left\_sector}(\state_l, \state_m) \land ...\right)  \right)  \lor \\
		& \left(\headon(\state_l, \state_m,\cdot)  \land \overtake(\state_m, \state_l,\cdot) \right) \\
		\overset{\mathrm{(b)}}{=}& \Big(\big(\mathrm{in\_front\_sector}(\state_l, \state_m) \land ... \big) \, \land \\
		&\big(\mathrm{in\_left\_sector}(\state_l, \state_m) \land ...\big)  \Big) \, \lor \\
		& \Big( \big( \mathrm{in\_front\_sector} (\state_l, \state_m) \, \land ... \big)  \, \land \\
		& \big( \mathrm{in\_behind\_sector}(\state_l, \state_m) \land ... \big)\Big)\\
		\overset{\mathrm{(c)}}{=}& \, \bot   \lor  \bot   \\
		=& \, \bot 
	\end{align*}
	(a) We distribute the disjunction in $\keep$ over the conjunction with $\headon$.
	
	\noindent (b) We insert the relevant parts of the predicates (see \cref{tab:predicates}).
	
	\noindent (c) For both parts of the disjunction, the vessels cannot be simultaneously in two sectors as in case (I).\smallskip	
\end{IEEEproof}

\subsection{Position Tracking Controller}\label{appendix:controller} 
We design a Lyapunov controller to track a desired position $\position_\mathrm{des}$ to realize the emergency maneuver control. This desired position is either the target position $\position_\mathrm{target}$ to be reached or generated based on the current position $\position_t$ and the desired position so that the vessel approximately maintains the desired velocity $v_\mathrm{desired}$. The Lyapunov function for turning rate $V_{\turingrate}$ and acceleration $V_{\nacceleration}$ are:
\begin{align*}
	V_{\turingrate} &= 1 - ([\cos(\orientation_t), \sin(\orientation_t)] \mathbf{w}_\mathrm{des}^T)^2,\\
	V_{\nacceleration} &= 0.5 \, (\position_\mathrm{des} - \position_t) (\position_\mathrm{des} - \position_t)^T ,
\end{align*}
with the desired orientation vector
\begin{equation*}
	 \mathbf{w}_\mathrm{des} = (\position_\mathrm{des} - \position_t) / \|\position_\mathrm{des} - \position_t\|_2.
\end{equation*}
With these Lyapunov functions, we obtain the control:
\begin{align*}
	\turingrate &= - \lambda_1 \, \frac{V_{\turingrate}}{- 2 \,  ([-\sin(\orientation), \cos(\orientation)] \mathbf{w}_\mathrm{des}^T) \, ([\cos(\orientation), \sin(\orientation)] \mathbf{w}_\mathrm{des}^T)},\\
	\nacceleration &= - \lambda_2 \, \frac{V_{\nacceleration}}{- (\position_\mathrm{des} - \position_t) [\nvelocity_t \cos(\orientation_t), \nvelocity_t \sin(\orientation_t)]^T}.
\end{align*}
If $V_{\turingrate}$ is larger than a threshold $\Delta_{V_{\turingrate}}$, then the acceleration control is set to zero so that the vessel only turns.
For our numerical evaluations, we use the following parameter values: $\nvelocity_\mathrm{desired}=\SI{6}{\meter \per \second}$, $\lambda_1=4$, $\lambda_2=0.04$, and $\Delta_{V_{\turingrate}} = 0.3$. 


\begin{IEEEbiography}[{\includegraphics[width=1in,height=1.25in,clip,keepaspectratio]{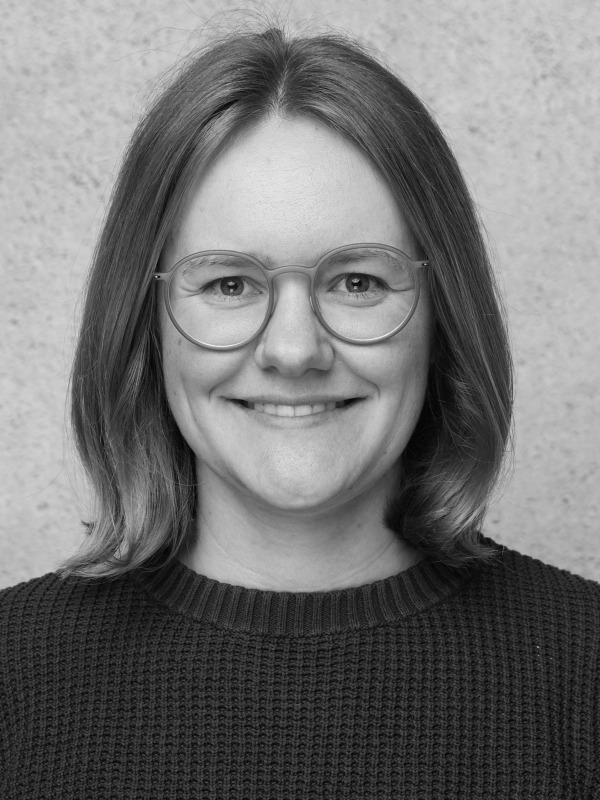}}]{Hanna Krasowski}
	is currently a Ph.D. candidate at the Technical University of Munich. She received her B.Sc. degree in mechanical engineering from Technical University of Darmstadt in 2017 and her M.Sc. degree in robotics, cognition and intelligence from Technical University of Munich in 2020.
	Her research interests include provably safe reinforcement learning and motion planning for cyber-physical systems.
\end{IEEEbiography}
\begin{IEEEbiography}[{\includegraphics[width=1in,height=1.25in,clip,keepaspectratio]{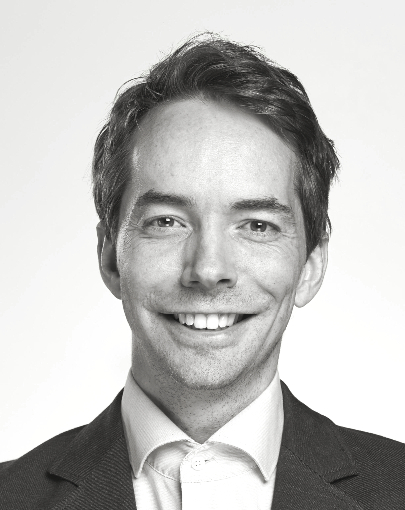}}]{Matthias Althoff}
	received the Diploma Engineering degree in mechanical engineering and the Ph.D. degree in electrical engineering from Technical University of Munich, Germany, in 2005 and 2010, respectively. He is currently an Associate Professor in computer science with Technical University of Munich, Germany. From 2010 to 2012 he was a Postdoctoral Researcher with Carnegie Mellon University, Pittsburgh, PA, USA, and from 2012 to 2013 an Assistant Professor with Technische Universität Ilmenau, Germany. His research interests include formal verification of continuous and hybrid systems, reachability analysis, planning algorithms, nonlinear control, automated vehicles, and power systems.
\end{IEEEbiography}

\begin{table*}[p]
	\vspace{0.2cm}
	\caption{Predicates for traffic rule specifications from \cite{Krasowski.2021} with adaptions for this work}
	\renewcommand{\arraystretch}{1.3}
	\centering
	\begin{tabular}{@{\extracolsep{4pt}}llll@{}}
		\toprule
		\multicolumn{1}{l}{\textbf{Predicate}} & \multicolumn{1}{l}{\textbf{Arguments}} & \multicolumn{1}{l}{\textbf{Definition}} &  \multicolumn{1}{l}{\textbf{Detects ...}} \\ \midrule 
		\multicolumn{4}{l}{\emph{Position and orientation predicates}} \\ \midrule
		$\mathrm{in\_front\_sector}$ & $\state_l, \state_m$ &  $\mathrm{in\_sector}(\state_l, \state_m,-\Delta_\mathrm{head\text{-}on}, \Delta_\mathrm{head\text{-}on})$ & \makecell[l]{relative position in \\ front sector} \smallskip  \\ 
		$\mathrm{in\_left\_sector}$ & $\state_l, \state_m$ &  $\mathrm{in\_sector}(\state_l, \state_m,\SI{-112.5}{\degree}, -\Delta_\mathrm{head\text{-}on})$ & \makecell[l]{relative position in \\ left sector} \smallskip  \\ 
		$\mathrm{in\_right\_sector}$ & $\state_l, \state_m$ &  $\mathrm{in\_sector}(\state_l, \state_m,\Delta_\mathrm{head\text{-}on}, \SI{112.5}{\degree})$ & \makecell[l]{relative position in \\ right sector} \smallskip  \\ 
		$\mathrm{in\_behind\_sector}$ & $\state_l, \state_m$ &  $\mathrm{in\_sector}(\state_l, \state_m,\SI{112.5}{\degree}, \SI{247.5}{\degree})$ & \makecell[l]{relative position in \\ behind sector} \smallskip \\  
		$\mathrm{orientation\_delta}$ & \makecell[l]{$\state_l, \state_m, $\\$ \Delta_{\mathrm{orient}}, c_\mathrm{o}$} & \makecell[l]{$\mathrm{mod}( \mathrm{proj}_\orientation (\state_m) - \mathrm{proj}_\orientation (\state_l) + c_\mathrm{o}, 2\pi)$ \\$\quad  \in [ \Delta_{\mathrm{orient}}, 2\pi-\Delta_{\mathrm{orient}}]$} & \makecell[l]{if relative orientation \\ is in defined range} \smallskip \\
		
		$ \mathrm{orientation\_towards\_right}$ & \makecell[l]{$\state_l, \state_m, $\\$ \Delta_{\mathrm{head\text{-}on}}$} & \makecell[l]{$\mathrm{mod}( \mathrm{proj}_\orientation (\state_m) - \mathrm{proj}_\orientation (\state_l), 2\pi)$ \\$\quad  \in [ - \pi + \Delta_{\mathrm{head\text{-}on}}, -\Delta_{\mathrm{head\text{-}on}}]$}& \makecell[l]{if relative orientation \\ of vessel $m$ is toward right} \smallskip \\ 
		
		$ \mathrm{orientation\_towards\_left}$ &  \makecell[l]{$\state_l, \state_m, $\\$ \Delta_{\mathrm{head\text{-}on}}$} & \makecell[l]{$\mathrm{mod}( \mathrm{proj}_\orientation (\state_m) - \mathrm{proj}_\orientation (\state_l), 2\pi)$ \\$\quad  \in [\Delta_{\mathrm{head\text{-}on}},  \pi - \Delta_{\mathrm{head\text{-}on}}]$}& \makecell[l]{if relative orientation \\ of vessel $m$ is toward right} \\ \midrule
		
		\multicolumn{4}{l}{\emph{Velocity predicates}} \\ \midrule
		$\mathrm{drives\_faster}$ & $\state_l, \state_m$ &  $\mathtt{proj}_\nvelocity (\state_l) > \mathtt{proj}_\nvelocity (\state_m)$ & \makecell[l]{if vessel $l$ is faster\\ than vessel $m$} \smallskip \\ 
		$\mathrm{safe\_speed}$ & $\state_l, \nvelocity_\mathrm{max}$ &  $0 \leq \mathtt{proj}_\nvelocity (\state_l) \leq \nvelocity_\mathrm{max}$ & safe speed of vessel $l$ \\  \midrule
		
		\multicolumn{4}{l}{\emph{General predicates}} \\ \midrule 
		$\mathrm{collision\_possible}$ & $\state_l, \state_m, t_\mathrm{horizon}$ &  \makecell[l]{$\mathcal{V}_l \in CC'(\state_l, \state_m) \land$ \\ $\|\mathbf{v}_l - \mathbf{v}_m\|_2 \geq \|\mathtt{proj}_{\position} (\state_l) - \mathtt{proj}_{\position} (\state_m)\|_2 / t_\mathrm{horizon}$} & \makecell[l]{if vessels $l$ and $m$ are \\ on a collision course} \smallskip \\ 
		
		$\mathrm{change\_course}$ & \makecell[l]{$\state_l, \mathcal{T}_l,$\\$ t_\mathrm{start}, \Delta_\mathrm{course}$} &  $| \sum_{t_i = t_\mathrm{start}}^{\mathtt{cl}(\mathcal{T}_l,\state_l)} \mathtt{proj}_{\turingrate}(\mathtt{state}(\mathcal{T}_l,t_i)) \, \Delta t | \geq \Delta_\mathrm{course}$ & \makecell[l]{if course has changed \\ significant since $t_\mathrm{start}$} \smallskip \\ 
		
		$\mathrm{turning\_to\_starbord}$ & $\state_l, \mathcal{T}_l,  t_\mathrm{start}$ &  \makecell[l]{$\mathrm{mod}\big(\mathtt{proj}_{\orientation}(\mathtt{state}(\mathcal{T}_l,\mathtt{cl}(\mathcal{T}_l,\state_l))) -$\\ $\mathrm{proj}_{\orientation}(\mathtt{state}(\mathcal{T}_l,t_\mathrm{start})), 2\pi\big) \in (\pi, 2\pi)$} & \makecell[l]{if course has changed to \\ starboard since $t_\mathrm{start}$} \smallskip \\
		
		$\mathrm{overtake}$ & \makecell[l]{$\state_l, \state_m, t_\mathrm{horizon}^\mathrm{check}$} &  \makecell[l]{$\mathrm{collision\_possible}(\state_l, \state_m, t_\mathrm{horizon}^\mathrm{check}) \, \land $ \\ $ \mathrm{in\_behind\_sector} (\state_m, \state_l) \, \land $ \\ $   \mathrm{drives\_faster}(\state_l, \state_m) \, \land$ \\ $\lnot \mathrm{orientation\_delta}(\state_l, \state_m, \SI{67.5}{\degree}, 0)$} & \makecell[l]{give-way vessel of overtaking \\ encounter situation}  \smallskip \\
		
		$\mathrm{maneuver\_overtake}$ & \makecell[l]{$\state_l, \state_m,$\\$ \mathcal{T}_l, t_\mathrm{horizon}^\mathrm{check},$\\$  \Delta_\mathrm{large\_turn}$} &  \makecell[l]{$\mathrm{change\_course}(\state_l, \mathcal{T}_n, \mathtt{t}_s(\mathrm{overtake}), \Delta_\mathrm{large\_turn})$} & \makecell[l]{correct maneuver of give-way \\ vessel in overtaking \\ encounter situation} \smallskip \\
		
		$\mathrm{head\_on}$ & \makecell[l]{$\state_l, \state_m, t_\mathrm{horizon}^\mathrm{check},$\\$\Delta_{\mathrm{head\text{-}on}}$} &  \makecell[l]{$\mathrm{collision\_possible}(\state_l, \state_m, t_\mathrm{horizon}^\mathrm{check}) \, \land $\\ $ \mathrm{in\_front\_sector} (\state_l, \state_m) \, \land $\\ $\lnot \mathrm{orientation\_delta}(\state_l, \state_m, \Delta_{\mathrm{head\text{-}on}}, \pi)$} & \makecell[l]{give-way vessel of head-on \\ encounter situation} \smallskip \\
		
		$\mathrm{maneuver\_head\_on}$ & \makecell[l]{$\state_l, \state_m, $\\$ \mathcal{T}_l, t_\mathrm{horizon}^\mathrm{check}, $\\$  \Delta_\mathrm{large\_turn},  \Delta_{\mathrm{head\text{-}on}}$} &  \makecell[l]{$\mathrm{change\_course}(\state_l, \mathcal{T}_n, \mathtt{t}_s(\mathrm{head\_on}), \Delta_\mathrm{large\_turn}) \land$ \\ $\mathrm{turning\_to\_starboard}(\state_l, \mathcal{T}_n, \mathtt{t}_s(\mathrm{head\_on}))$} & \makecell[l]{correct maneuver of give-way \\ vessel in head-on \\ encounter situation} \smallskip \\
		
		$\mathrm{crossing}$ & \makecell[l]{$\state_l, \state_m, t_\mathrm{horizon}^\mathrm{check},$\\$\Delta_{\mathrm{head\text{-}on}}$} &  \makecell[l]{$\mathrm{collision\_possible}(\state_l, \state_m, t_\mathrm{horizon}^\mathrm{check}) \land $\\$ \mathrm{in\_right\_sector} (\state_l, \state_m) \land $ \\$\mathrm{orientation\_towards\_left} (\state_l, \state_m, \Delta_{\mathrm{head\text{-}on}})$} & \makecell[l]{give-way vessel of crossing \\ encounter situation} \smallskip \\
		$\mathrm{maneuver\_crossing}$ & \makecell[l]{$\state_l, \state_m, $\\$ \mathcal{T}_l, t_\mathrm{horizon}^\mathrm{check}, $\\$ \Delta_\mathrm{large\_turn},   \Delta_{\mathrm{head\text{-}on}}$} &  \makecell[l]{$\mathrm{change\_course}(\state_l, \mathcal{T}_n, \mathtt{t}_s(\mathrm{crossing}), \Delta_\mathrm{large\_turn}) \land$ \\ $\mathrm{turning\_to\_starboard}(\state_l, \mathcal{T}_n, \mathtt{t}_s(\mathrm{crossing}))$} & \makecell[l]{correct maneuver of give-way \\ vessel in crossing \\ encounter situation}  \smallskip \\
		
		$\mathrm{keep}$ & \makecell[l]{$\state_l, \state_m, t_\mathrm{horizon}^\mathrm{check},$\\$\Delta_{\mathrm{head\text{-}on}}$} &  \makecell[l]{$\big(\mathrm{collision\_possible}(\state_l, \state_m, t_\mathrm{horizon}^\mathrm{check}) \land $\\$ \mathrm{in\_left\_sector} (\state_l, \state_m) \land $ \\$ \mathrm{orientation\_towards\_right} (\state_l, \state_m, \Delta_{\mathrm{head\text{-}on}}) \big) \lor$ \\ $\mathrm{overtake}(\state_m, \state_l, t_\mathrm{horizon}^\mathrm{check})$} & \makecell[l]{stand-on vessel} \smallskip \\
		
		$\mathrm{no\_turning}$ & $\state_l,  \mathcal{T}_l, \Delta_\mathrm{no\_turn}$ &  \makecell[l]{$\lnot \mathrm{change\_course}(x_n, \mathcal{T}_n, \mathtt{t}_s(\mathrm{keep}), \Delta_\mathrm{no\_turn})$} & \makecell[l]{correct stand-on maneuver} \smallskip \\   \bottomrule 
	\end{tabular}
	\label{tab:predicates}
\end{table*} 

\end{appendices}

\end{document}